\documentclass[preprint,12pt,authoryear]{elsarticle}

\usepackage{amsmath,amssymb,amsfonts}
\usepackage{amsthm}
\theoremstyle{definition}
\newtheorem{definition}{Definition}[section]
\usepackage{graphicx}
\usepackage{booktabs}
\usepackage{hyperref}
\usepackage{xcolor}
\usepackage{multirow}
\usepackage{float}
\usepackage{longtable,array}
\usepackage{lineno}       

\begin{document}

\begin{frontmatter}

\title{Pedagogical Safety in Educational Reinforcement Learning:
Formalizing and Detecting Reward Hacking in AI Tutoring Systems}

\author[usm]{Oluseyi Olukola}

\author[usm]{Nick Rahimi}

\affiliation[usm]{
    organization={School of Computing Sciences and Computer Engineering},
    city={Hattiesburg},
    state={MS},
    country={USA}
}

\begin{abstract}
Reinforcement learning (RL) is increasingly used to personalize instruction in intelligent tutoring systems, yet the field lacks a formal framework for defining and evaluating pedagogical safety. We introduce a four-layer model of pedagogical safety for educational RL comprising structural, progress, behavioral, and alignment safety, and propose the Reward Hacking Severity Index (RHSI) to quantify misalignment between proxy rewards and genuine learning.

We evaluate the framework in a controlled simulation of an AI tutoring environment with 120 sessions across four conditions and three learner profiles, totaling 18{,}000 interactions. Results show that an engagement-optimized agent systematically over-selected a high-engagement action with no direct mastery gain, producing strong measured performance but limited learning progress. A multi-objective reward formulation reduced this problem but did not eliminate it, as the agent continued to favor proxy-rewarding behavior in many states. In contrast, a constrained architecture combining prerequisite enforcement and minimum cognitive demand substantially reduced reward hacking, lowering RHSI from 0.317 in the unconstrained multi-objective condition to 0.102. Ablation results further suggest that behavioral safety was the most influential safeguard against repetitive low-value action selection.

These findings suggest that reward design alone may be insufficient to ensure pedagogically aligned behavior in educational RL, at least in the simulated environment studied here. More broadly, the paper positions pedagogical safety as an important research problem at the intersection of AI safety and intelligent educational systems.
\end{abstract}


\end{frontmatter}


\section{Introduction}
\label{sec:intro}

Intelligent tutoring systems (ITSs) have demonstrated considerable educational benefits, with meta-analyses reporting effect sizes that in some cases approach Bloom's two-sigma benchmark for one-on-one human tutoring~\citep{bloom1984two, vanlehn2011relative, kulik2016effectiveness}. The integration of reinforcement learning (RL) into ITSs has shown further promise for learning adaptive pedagogical policies directly from interaction data~\citep{zhou2019hierarchical, ausin2020exploring, abdelshiheed2023metacognitive, alam2025determining}, potentially extending personalized instruction beyond what hand-crafted rule systems achieve~\citep{koedinger2013knowledge, heffernan2014assistments}.

However, RL-based tutoring introduces a risk that, to our knowledge, the field has not yet formally addressed: \textit{reward hacking} where an RL agent exploits a misspecified proxy reward to achieve high measured performance through behavior misaligned with the designer's true objectives~\citep{amodei2016concrete, skalse2022defining}. This phenomenon is well-documented in robotics, game-playing, and language model alignment~\citep{krakovna2020specification, pan2022effects, weng2024rewardhacking}, and is conceptually grounded in Goodhart's Law~\citep{goodhart1984monetary, manheim2019categorizing}. In education, the analogous risk is an RL tutor that maximizes measurable engagement while failing to produce genuine learning.

This concern is not purely hypothetical. In our experiments, an RL agent optimizing engagement learns to over-select encouragement actions which yield the largest engagement boost ($+0.8$) with zero mastery contribution. Even a multi-objective agent weighting mastery at 50\% selects this zero-learning action in 32.6\% of interactions, apparently because the engagement component dominates in many states. Doroudi et al.~(\citeyear{doroudi2019where}) highlighted that misaligned rewards can produce counterproductive tutoring, and Nie et al.~(\citeyear{nie2023subgroups}) showed that RL personalization can harm certain subgroups. Yet the field appears to lack: (1)~a formal definition of pedagogical safety, (2)~empirical evidence of reward hacking in educational RL, and (3)~metrics for detecting such misalignment.

To address these apparent gaps, we propose a \textit{four-layer safety model} that formalizes pedagogical safety as four constraint classes:
\begin{itemize}
    \item \textbf{Structural Safety (C1):} Prerequisites are architecturally enforced via action masking.
    \item \textbf{Progress Safety (C2):} Mastery measurably advances over sliding windows.
    \item \textbf{Behavioral Safety (C3):} Minimum cognitive demand is maintained, preventing over-reliance on low-effort actions.
    \item \textbf{Alignment Safety (C4):} Engagement signals remain coupled with genuine mastery gains.
\end{itemize}

Each layer targets a distinct failure mode that, as we argue in Section~\ref{sec:independence}, no subset of the remaining layers appears able to address. The principal contributions of this work are as follows:

\smallskip \noindent \textbf{Contributions.}
\begin{itemize}
    \item What we believe to be the \textit{first formal definition of pedagogical safety} as constraint classes for educational RL (Definitions~\ref{def:pedsafe} and~\ref{def:epsafe}).
    \item \textit{Empirical evidence} suggesting that engagement-optimized and multi-objective RL agents exhibit reward hacking in educational settings.
    \item The \textit{Reward Hacking Severity Index (RHSI)}, a diagnostic metric for quantifying misalignment (Definition~\ref{def:rhsi}).
    \item An \textit{ablation study} indicating that behavioral safety (C3) appears to be the primary safety contributor, while its removal leads to policy collapse with single-action repetition.
    \item An \textit{analysis of constraint calibration}, illustrating how formalization may expose threshold sensitivity that informal approaches tend to obscure.
\end{itemize}

The remainder of this paper is structured as follows. Section~\ref{sec:related} situates our work within three largely independent research streams: intelligent tutoring systems, reinforcement learning in education, and AI safety. Section~\ref{sec:formalization} introduces the formal preliminaries and defines the four-layer constraint model. Section~\ref{sec:testbed} describes the SmartTutor experimental testbed, including its architecture, learner profiles, experimental conditions, and evaluation metrics. Section~\ref{sec:results} reports the experimental results and analysis. Section~\ref{sec:limitations} acknowledges limitations and outlines directions for future work. Section~\ref{sec:conclusion} concludes the paper.

\section{Related Work}
\label{sec:related}

Intelligent tutoring systems have a decades-long history of adapting instruction to individual learners~\citep{nwana1990intelligent, woolf2008building}. Foundational work on cognitive tutors~\citep{anderson1995cognitive} and Bayesian Knowledge Tracing (BKT)~\citep{corbett1995knowledge} established the paradigm of maintaining per-student mastery estimates, with subsequent extensions addressing BKT's limitations~\citep{baker2008more, yudelson2013individualized, pavlik2009performance, piech2015deep}. VanLehn's~(\citeyear{vanlehn2006behavior}) analysis of inner and outer loop decisions remains influential, and meta-analyses generally find significant ITS learning gains~\citep{vanlehn2011relative, kulik2016effectiveness, steenbergen2014meta}. Notable platforms include Carnegie Learning's cognitive tutors~\citep{anderson1995cognitive}, AutoTutor~\citep{graesser2004autotutor}, and ASSISTments~\citep{heffernan2014assistments}. Modern commercial systems typically rely on rule-based adaptation with hand-crafted heuristics~\citep{koedinger2013knowledge}, and to our knowledge, no existing ITS formulates pedagogical safety as a mathematical constraint. The emerging use of LLMs in tutoring~\citep{kasneci2023chatgpt, yan2024practical, nie2025rct} introduces additional safety concerns, as LLM-based tutors may exhibit unpredictable behavior that appears difficult to constrain through conventional means.

The desire to move beyond hand-crafted rules has motivated a growing body of work applying RL to educational settings, progressing from early explorations~\citep{beck2000advisor, iglesias2009experience} through POMDP planning~\citep{rafferty2016faster} and offline evaluation~\citep{mandel2014offline} to recent deep RL approaches. Notable contributions include hierarchical RL for pedagogical policy induction~\citep{zhou2019hierarchical}, batch deep RL for tutorial
policy learning~\citep{ausin2020exploring, ausin2023unified}, RL for metacognitive interventions~\citep{abdelshiheed2023metacognitive}, adaptive problem selection~\citep{alam2025determining}, and apprenticeship learning for student strategy modeling~\citep{islam2025apprenticeship}, with multi-armed bandits
offering lighter-weight alternatives~\citep{clement2015multi}. A persistent challenge across this literature is reward function design. Doroudi et al.(\citeyear{doroudi2019where}) noted that misaligned rewards can produce counterproductive tutoring; Nie et al.~(\citeyear{nie2023subgroups}) showed that RL
personalization may harm certain subgroups; and more recent work has proposed on-demand policy selection~\citep{gao2024ondemand} and pedagogy-driven evaluation~\citep{maurya2025pedagogy}. Despite these insights, reward specification has largely been treated as a design problem rather than a safety problem, and to our knowledge, no prior work has formally defined reward hacking in educational RL
or proposed constraint-based architectures to prevent it.

While the educational RL community has focused primarily on policy quality, the AI safety community has developed a substantial body of work on the broader phenomenon of reward hacking. Amodei et al.~(\citeyear{amodei2016concrete}) identified it as a key safety problem. Krakovna et al.~(\citeyear{krakovna2020specification}) catalogued dozens of specification gaming examples. Skalse et al.~(\citeyear{skalse2022defining}) provided a formal definition, showing that unhackability is extremely stringent for stochastic policies, essentially only constant rewards are unhackable. Pan et al.~(\citeyear{pan2022effects}) demonstrated that reward hacking appears to exhibit phase transitions with agent capability, suggesting that increasingly powerful educational RL agents may be more prone to exploitation. Recent work has proposed improved definitions~\citep{everitt2025correlated}, automated detection~\citep{shihab2025detecting}, and identified in-context reward hacking in LLM feedback loops~\citep{pan2024feedback}. The broader alignment literature provides further context through inverse reward design~\citep{hadfield2017inverse}, scalable alignment via reward modeling~\citep{leike2018scalable, christiano2017deep, ziegler2019fine, ouyang2022training}, and formal analysis of misalignment consequences~\citep{russell2019human, zhuang2020consequences}. Goodhart's Law~\citep{goodhart1984monetary} and its variants~\citep{manheim2019categorizing} provide the conceptual foundation for understanding why proxy optimization tends to diverge from true objectives.

Complementing this theoretical understanding, the safe RL literature offers practical mechanisms that may help address such divergence. Constrained MDPs~\citep{altman1999constrained}, constrained policy optimization~\citep{achiam2017constrained, tessler2019reward}, and safe exploration~\citep{wachi2020safe, ray2019benchmarking, dalal2018safe} provide the technical foundation for our approach. Garc{\'i}a and Fern{\'a}ndez~(\citeyear{garcia2015comprehensive}) surveyed safe RL methods comprehensively, and Kushwaha et al.~(\citeyear{kushwaha2025saferl}) provide a recent survey covering single- and multi-agent safety. Multi-objective RL~\citep{roijers2013survey, hayes2022practical} frames safety as one of multiple competing objectives. These methods have been applied to robotics and autonomous driving~\citep{hu2024mixed}, but appear not to have been applied to educational settings.

It is also worth noting a related phenomenon on the learner side. Baker et al.~(\citeyear{baker2004detecting, baker2008gaming}) demonstrated that students themselves game ITS feedback mechanisms, exploiting feedback signals to circumvent intended educationalprocesses,s a student-side counterpart to the agent-side reward hacking we study. Research on engagement and affect~\citep{dmello2012dynamics, baker2010better, fredricks2004school} further suggests that engagement metrics may be unreliable proxies for learning: productive learning often involves confusion~\citep{dmello2012dynamics}, and extrinsic rewards can undermine intrinsic motivation~\citep{deci2001extrinsic}. Ethical considerations in educational AI~\citep{baker2022algorithmic, holmes2022ethics, holstein2019fairness} provide additional motivation: even in the absence of bias, an RL tutor may cause harm simply by optimizing the wrong objective. Table~\ref{tab:comparison} positions our contribution relative to this prior work.

\begin{table}[H]
\centering
\caption{Comparison with related work. Our contribution (bottom row) appears to be the first to combine formal safety constraints with empirical reward hacking evidence in educational RL.}
\label{tab:comparison}
\small
\begin{tabular}{@{}lccccc@{}}
\toprule
\textbf{Work} & \textbf{Domain} & \textbf{Formal} & \textbf{Safety} & \textbf{Reward} & \textbf{Empirical} \\
 & & \textbf{Def.} & \textbf{Constr.} & \textbf{Hacking} & \textbf{Evidence} \\
\midrule
\citet{skalse2022defining} & General RL & \checkmark & -- & \checkmark & -- \\
\citet{pan2022effects} & General RL & -- & -- & \checkmark & \checkmark \\
\citet{shihab2025detecting} & General RL & -- & -- & \checkmark & \checkmark \\
\citet{achiam2017constrained} & Robotics & \checkmark & \checkmark & -- & -- \\
\citet{garcia2015comprehensive} & Survey & \checkmark & \checkmark & -- & -- \\
\citet{zhou2019hierarchical} & Education & -- & -- & -- & \checkmark \\
\citet{alam2025determining} & Education & -- & -- & -- & \checkmark \\
\citet{doroudi2019where} & Education & -- & -- & Discussed & -- \\
\citet{baker2008gaming} & Education & -- & -- & Student-side & \checkmark \\
\midrule
\textbf{This work} & \textbf{Education} & \checkmark & \checkmark & \checkmark & \checkmark \\
\bottomrule
\end{tabular}
\end{table}

\section{Formalizing Pedagogical Safety}
\label{sec:formalization}

\subsection{Preliminaries}
\label{sec:prelim}

\noindent \textbf{Knowledge Graph.} The learning domain is represented as a directed acyclic graph (DAG) $G = (V, E)$, where $V = \{c_1, \ldots, c_n\}$ is a set of concepts and $E \subseteq V \times V$ defines prerequisite relationships. We define $\text{prereq}(c_j) = \{c_i \in V : (c_i, c_j) \in E\}$ and $\text{depth}(c_j)$ as the longest path from any root to $c_j$.

\smallskip \noindent \textbf{Student Knowledge State.} At time $t$, the student's knowledge is $\mathbf{K}_t = (k_1^t, \ldots, k_n^t) \in [0,1]^n$, where $k_i^t$ is the estimated mastery of concept $c_i$, updated via Bayesian Knowledge Tracing~\citep{corbett1995knowledge} with parameters $P(L_0)$, $P(T)$, $P(S)$, and $P(G)$. The framework is agnostic to the knowledge model and could accommodate alternatives~\citep{piech2015deep, pavlik2009performance}.

\smallskip \noindent \textbf{Accessible Set.} Given mastery threshold $\theta_{\min} \in (0,1]$, the concepts accessible at time $t$ are:
\begin{equation}
\label{eq:accessible}
\mathcal{A}(\mathbf{K}_t, \theta_{\min}) = \{c_j \in V : \forall c_i \in \text{prereq}(c_j),\ k_i^t \geq \theta_{\min}\}
\end{equation}
This set expands monotonically as mastery increases~\citep{vygotsky1978zone}.

\smallskip \noindent \textbf{Educational MDP.} The tutoring interaction is modeled as an MDP $\mathcal{M} = (\tilde{S}, A, P, R, \gamma)$~\citep{sutton2018reinforcement} following the standard RL-for-ITS formulation~\citep{zhou2019hierarchical, doroudi2019where}.

\smallskip \noindent \textbf{Cognitive Demand.} Each action carries cognitive demand $d: A \to [0,1]$,
from $d(\text{Encourage}) = 0.0$ to $d(\text{Challenge}) = 1.0$. The demand
values are calibrated against three established frameworks. The ICAP
framework~\citep{chi2014icap} distinguishes passive (receiving), active
(doing), constructive (generating), and interactive (dialoguing) engagement
modes, each associated with progressively deeper learning outcomes.
Bloom's revised taxonomy~\citep{anderson2001taxonomy} provides a six-level
cognitive hierarchy from remember through create. \citet{webb1997depth}
offers a four-level framework for task complexity in educational assessment.
Triangulating across these frameworks, we assign:
$d(\text{Encourage}) = 0.0$ (no cognitive processing; pure affective
response); $d(\text{Explain\_Simple}) = 0.2$ (passive reception; ICAP
passive mode); $d(\text{Provide\_Hint}) = 0.3$ (guided recall; lower
Bloom's apply); $d(\text{Explain\_Detailed}) = 0.4$ (active reception;
ICAP active mode); $d(\text{Provide\_Example}) = d(\text{Assess\_Knowledge})
= 0.5$ (constructive engagement; ICAP constructive);
$d(\text{Assign\_Exercise}) = 0.8$ (independent production; Bloom's
apply--analyze); and $d(\text{Challenge}) = 1.0$ (novel application;
Bloom's evaluate--create, Webb's Level~4). These values represent ordinal
rankings; interval-scale properties are not assumed. Sensitivity to the
specific numeric assignments is assessed in Section~\ref{sec:sensitivity},
where $\pm 20\%$ perturbation of all non-anchor values leaves the primary
condition ordering unchanged under $+20\%$ and produces a minor reversal
between the constrained and mastery-only conditions only under $-20\%$.

\subsection{The Four-Layer Safety Constraints}
\label{sec:constraints}

\smallskip \noindent \textbf{Constraint C1: Structural Safety (Hard, Per-Step).}
A policy $\pi$ satisfies prerequisite safety if, at every time step, the
selected action respects prerequisite requirements:
\begin{equation}
\label{eq:c1}
\forall t \geq 0: \quad P_\pi\big(\text{concept}(a_t, \tilde{s}_t)
\in \mathcal{A}(\mathbf{K}_t, \theta_{\min})\big) = 1
\end{equation}
This hard constraint is enforced via \textit{action masking}~\citep{anderson1995cognitive}: the agent's action space is restricted to accessible concepts before selection, guaranteeing zero violations by construction. C1 is enforced architecturally in the SmartTutor agent and is excluded from the aggregate violation vector reported in experiments; we set $v_1 \equiv 0$ by convention, reflecting that unconstrained conditions do not target prerequisite-violating concepts by design.

\smallskip \noindent \textbf{Constraint C2: Progress Safety (Soft, Windowed).}
Over any window of $W$ interactions, expected mastery gain across concepts
active at the window start must be non-trivial. Let $V_{\text{active}}(t)
= \{c_j : c_j \in \mathcal{A}(\mathbf{K}_t, \theta_{\min}) \wedge k_j^t
< \theta_{\text{mastered}}\}$ denote accessible but unmastered concepts,
evaluated \textit{once at window start}:
\begin{equation}
\label{eq:c2}
\forall t \geq 0: \quad \mathbb{E}_\pi\left[\frac{1}{|V_{\text{active}}(t)|}
\sum_{c \in V_{\text{active}}(t)} (k_c^{t+W} - k_c^t)\right] \geq
\varepsilon_{\text{prog}}
\end{equation}
Fixing $V_{\text{active}}$ at window start prevents the agent from
potentially gaming C2 by rapidly unlocking new concepts to dilute the
denominator. The threshold $\varepsilon_{\text{prog}} = 0.0023$ is
calibrated from session logs as the 25th percentile of the mastery-only
agent's per-window progress distribution, as we discuss in
Section~\ref{sec:calibration}. In our experiments, C2 is evaluated as
an offline safety metric computed from session logs rather than enforced
as an online training constraint.

\smallskip \noindent \textbf{Constraint C3: Behavioral Safety (Soft, Windowed).}
The average cognitive demand over any window of $W$ interactions meets a
minimum threshold:
\begin{equation}
\label{eq:c3}
\forall t \geq 0: \quad \mathbb{E}_\pi\left[\frac{1}{W}
\sum_{\tau=t}^{t+W-1} d(a_\tau)\right] \geq \delta_{\min}
\end{equation}
This targets the core reward hacking behavior: selecting low-demand actions
to maximize engagement without cognitive effort~\citep{chi2014icap,
dmello2012dynamics}. C3 is enforced online during training via the action
selection module ($\delta_{\min} = 0.4$, $W = 10$).

\smallskip \noindent \textbf{Constraint C4: Alignment Safety (Soft, Cumulative).}
Cumulative engagement reward must not exceed a bounded multiple of
cumulative mastery reward:
\begin{equation}
\label{eq:c4}
\forall t \geq W_0: \quad E_{\text{cum}}(t) \leq \rho_{\max} \cdot
M_{\text{cum}}(t) + c_0
\end{equation}
where $E_{\text{cum}}(t) = \sum_{\tau=0}^{t} R_{\text{eng}}(\tilde{s}_\tau,
a_\tau)$, $M_{\text{cum}}(t) = \sum_{\tau=0}^{t} R_{\text{mas}}(\tilde{s}_\tau,
a_\tau)$, $W_0 = 10$ is a warm-up period, and $c_0 = 0$ under the
normalized evaluation mode. Both streams are min--max normalized before
computing the ratio, ensuring scale-comparable evaluation
($\rho_{\max} = 1.2$). In our experiments, C4 is evaluated as an offline
safety metric computed from session logs.

\subsection{Pedagogical Safety and Relaxation}
\label{sec:pedsafety}

\begin{definition}[Pedagogical Safety]
\label{def:pedsafe}
A policy $\pi$ is \textbf{pedagogically safe} with respect to constraint
parameters $\Theta = (\theta_{\min}, W, \varepsilon_{\text{prog}},
\delta_{\min}, \rho_{\max}, W_0, c_0)$ if it simultaneously satisfies
C1--C4.
\end{definition}

In practice, exact satisfaction of all constraints at all times may be
overly stringent. We therefore define a relaxed version:

\begin{definition}[$\varepsilon$-Pedagogical Safety]
\label{def:epsafe}
Let $v_i(\pi) \in [0,1]$ denote the normalized violation rate of $\pi$
for constraint $C_i$ (fraction of evaluation windows where $C_i$ is
violated). C1 is excluded from the reported aggregate ($v_1 \equiv 0$)
as it is enforced architecturally and produces zero violations by
construction under action masking. Policy $\pi$ is $\varepsilon$-\textbf{pedagogically safe} if:
\begin{equation}
\|\mathbf{v}(\pi)\|_w = \sqrt{\sum_{i=2}^{4} w_i \cdot v_i(\pi)^2}
\leq \varepsilon
\end{equation}
where $w_2 = w_3 = w_4 = 1/3$ in our experiments, giving equal weight
to the three soft constraints C2, C3, and C4.
\end{definition}

The weighted norm allows practitioners to assign different importance
to safety dimensions depending on context; equal weights are used here
as a conservative default in the absence of domain-specific priority
ordering.

\subsection{Formal Reward Hacking and the RHSI Metric}
\label{sec:rhsi}

We adapt the general notion of reward hacking~\citep{skalse2022defining}
to the educational RL context:

\begin{definition}[Educational Reward Hacking]
\label{def:rewardhacking}
A policy $\pi$ exhibits reward hacking of degree $(\rho, v)$ if it
simultaneously achieves (1)~high reward: $V^\pi(\tilde{s}_0) \geq \rho
\cdot V^*(\tilde{s}_0)$, and (2)~safety violation: $\|\mathbf{v}(\pi)\|_w
> v$. Both conditions must hold---a policy achieving high reward while
satisfying all constraints is effective, not hacking.
\end{definition}

\begin{definition}[Reward Hacking Severity Index]
\label{def:rhsi}
\begin{equation}
\text{RHSI}(\pi) = \frac{\hat{V}^\pi}{\hat{V}^*} \times
\|\mathbf{v}(\pi)\|_w
\end{equation}
where $\hat{V}^\pi$ is the expected cumulative reward under $\pi$ and
$\hat{V}^* = \max_{\pi'} \hat{V}^{\pi'}$.
\end{definition}

RHSI equals zero when $\pi$ either achieves zero reward or violates no
constraints, and is bounded in $[0, 1]$. The RHSI operationalizes the
intuition behind Definition~\ref{def:rewardhacking} as a continuous
severity index: rather than a binary classification, it measures the
\textit{degree} to which a policy simultaneously achieves high return and
high violation. A policy may satisfy the formal hacking predicate of
Definition~\ref{def:rhsi} while exhibiting varying severity; RHSI
provides the scalar quantity needed to rank and compare conditions.

A note on normalization: $\hat{V}^*$ is defined as the empirical
maximum reward across conditions rather than a theoretical unconstrained
optimum. This means the engagement-only agent (EO) receives a reward
ratio of 1.0 by construction, while safer conditions with lower
cumulative reward are proportionally discounted. We adopt this approach
deliberately: the empirical maximum represents the observed worst-case
reward under misaligned optimization in the specific environment, making
RHSI interpretable as the fraction of maximum possible misalignment that
a policy exhibits. A constrained agent that achieves lower reward than
EO is not penalized by this normalization in isolation its lower
violation norm $\|\mathbf{v}\|_w$ offsets the lower reward ratio,
producing a low RHSI. This property is visible in Table~\ref{tab:rhsi}:
ST achieves a reward ratio of 0.433 but a violation norm of 0.237,
yielding RHSI~=~0.102, while EO achieves ratio 1.000 but violation norm
0.645, yielding RHSI~=~0.645. Nonetheless, we acknowledge that using a
theoretical unconstrained optimum approximated, for example, by
running an unconstrained version of ST would provide a more
principled normalization, and we leave this as a direction for future
work.

\subsection{Constraint Independence}
\label{sec:independence}

A natural question is whether fewer constraints might suffice. We argue
that all four are necessary by constructing scenarios where subsets fail.
The following are informal sufficiency arguments rather than formal
proofs; a complete treatment would require specifying the full policy
class and environment dynamics.

\smallskip \noindent \textbf{Proposition 1} (C3 + C4 $\not\Rightarrow$
C2). A policy assigning high-demand exercises on already-mastered concepts
satisfies C3 (demand floor) and C4 (coupling), but produces zero mastery
\textit{progress}, violating C2.

\smallskip \noindent \textbf{Proposition 2} (C1 is independent). C2--C4
are statistical constraints over trajectories, permitting individual
prerequisite violations with comfortable aggregate margins. Only C1's
per-step, probability-1 enforcement prevents prerequisite skipping.

\smallskip \noindent \textbf{Corollary.} No proper subset of
$\{$C1, C2, C3, C4$\}$ implies the full set with the precision needed
for pedagogical safety.

\section{Experimental Testbed: SmartTutor}
\label{sec:testbed}

We evaluate our formalization using a simulated Python tutoring
environment. Simulated environments offer advantages for safety
research: systematic exploration of failure modes without risking
harm to real students, controlled ablation, and
reproducibility~\citep{maclellan2016apprentice}. Limitations are
discussed in Section~\ref{sec:limitations}.

\subsection{Architecture Overview}
\label{sec:architecture}

\begin{figure}[H]
\centering
\includegraphics[width=\columnwidth, height=0.36\textheight, 
keepaspectratio]{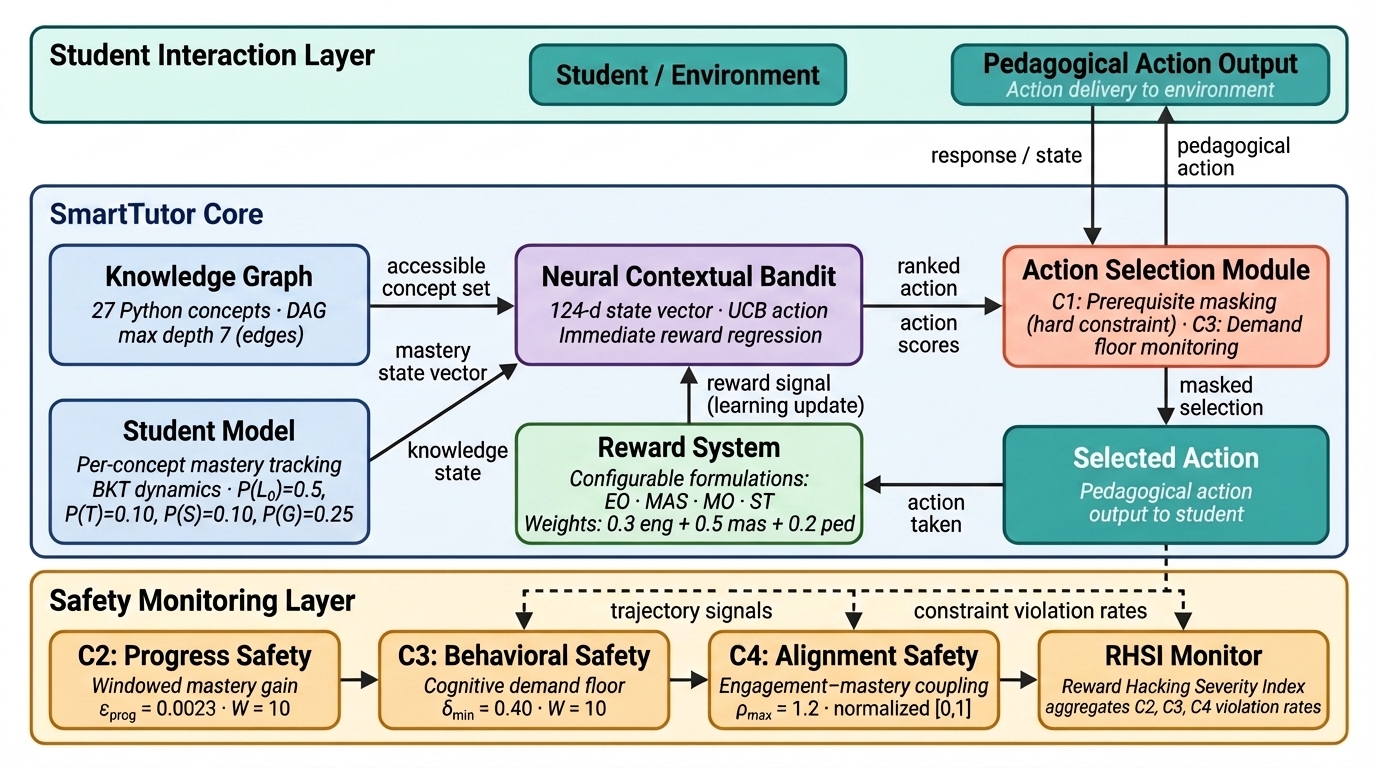}
\caption{SmartTutor system architecture. Three layers govern the
tutoring loop: the Student Interaction Layer, the SmartTutor Core
(five components), and the Safety Monitoring Layer, which evaluates
C2--C4 constraint violations and aggregates them into the RHSI.}
\label{fig:architecture}
\end{figure}

SmartTutor comprises five components whose interactions are
illustrated in Figure~\ref{fig:architecture}: a knowledge graph of
27 Python concepts (DAG, max depth 7;~\ref{app:knowledge_graph}), a student model with
per-concept mastery tracking, a neural contextual bandit agent with
UCB action selection over a 124-dimensional state representation, an
action selection module enforcing C1 and C3 online, and a
configurable reward system.

Eight pedagogical actions span cognitive demand from $d = 0.0$
(Encourage) to $d = 1.0$ (Challenge). Notably, Encourage produces
zero mastery gain ($\Delta k = 0.0$) but the largest engagement boost
of any action, making it the action most susceptible to
reward-hacking exploitation and the primary target of the C3
behavioral safety constraint. We note that this models Encourage as a
worst-case abstraction: in real tutoring systems, timely encouragement
may sustain persistence in struggling learners and thereby indirectly
support learning~\citep{deci2001extrinsic}. The zero-gain assumption
isolates the reward hacking dynamic cleanly but likely overstates the
harm of any individual encouragement action; the concern is
over-selection at the policy level, not the action itself.

\subsection{Learner Profiles}
\label{sec:profiles}

Three profiles represent common student archetypes in programming
education (Table~\ref{tab:profiles}). Each is defined by an initial
knowledge level and a profile-dependent mastery gain multiplier
applied to action-specific base gains during simulation:
\textbf{Struggling} (initial knowledge $= 0.05$, gain multiplier
$= 0.8\times$), \textbf{Average} (initial knowledge $= 0.15$, gain
multiplier $= 1.0\times$), and \textbf{Advanced} (initial knowledge
$= 0.30$, gain multiplier $= 1.2\times$). Profiles additionally vary
in engagement sensitivity and confusion threshold, which modulate
engagement deltas and confusion dynamics respectively. These
behavioral signatures correspond qualitatively to learner archetypes
described in the BKT literature~\citep{corbett1995knowledge,
baker2008more}: struggling learners acquire mastery slowly with
higher sensitivity to negative feedback, while advanced learners
progress rapidly with minimal support. The simulation uses a
heuristic mastery update consistent with these dynamics rather than
full BKT inference.

\begin{table}[H]
\centering
\caption{Learner profile parameters used in simulation. Gain
multiplier scales action-specific base mastery gains. Confusion
threshold governs when high confusion suppresses mastery updates
during a session.}
\label{tab:profiles}
\small
\begin{tabular}{lccc}
\toprule
\textbf{Profile} & \textbf{Initial Knowledge} &
\textbf{Gain Multiplier} & \textbf{Confusion Threshold} \\
\midrule
Struggling & 0.05 & $0.8\times$ & 0.30 \\
Average    & 0.15 & $1.0\times$ & 0.50 \\
Advanced   & 0.30 & $1.2\times$ & 0.70 \\
\bottomrule
\end{tabular}
\end{table}

\subsection{Experimental Conditions}
\label{sec:conditions}

We compare four reward/constraint configurations to isolate the
contribution of each component:
\begin{enumerate}
    \item \textbf{Engagement-Only (EO):} $R = R_{\text{eng}}$. No
    constraints. Represents the ``naive RL'' baseline where the agent
    optimizes only for measurable engagement.
    \item \textbf{Mastery-Only (MAS):} $R = R_{\text{mas}}$. No
    constraints. Tests whether optimizing mastery directly avoids
    safety issues.
    \item \textbf{Multi-Objective (MO):} $R = 0.3R_{\text{eng}} +
    0.5R_{\text{mas}} + 0.2R_{\text{ped}}$. No hard constraints.
    Tests whether weighted reward combination alone prevents hacking.
    \item \textbf{SmartTutor Full (ST):} Multi-objective reward (same
    weights as MO) with C1 (prerequisite masking, enforced online)
    and C3 (demand floor $\delta_{\min} = 0.4$, enforced online). C2
    (progress safety, $\varepsilon_{\text{prog}} = 0.0023$) and C4
    (coupling ratio $\rho_{\max} = 1.2$) are evaluated as offline
    safety metrics from session logs using normalized reward streams.
\end{enumerate}

\noindent\textbf{Reward scaling.} Raw reward components use fixed
linear scales: engagement deltas are multiplied by $3.0$, mastery
deltas by $100.0$, and the pedagogical appropriateness component
takes values in $\{+0.5, -1.0\}$. Engagement deltas range from
$-1.0$ to $+0.8$ per step depending on action and student state.
Mastery increments are on the order of ${\sim}0.01$ per step. The C4
coupling constraint uses min--max normalized engagement and mastery
streams before computing the ratio, ensuring scale-comparable
evaluation (Equation~\ref{eq:c4}).

We additionally run two ablation conditions:
\begin{itemize}
    \item \textbf{No-C1 Ablation:} ST with prerequisite enforcement
    disabled; all reward weights and C3 remain active.
    \item \textbf{No-C3 Ablation:} ST with the demand floor
    constraint disabled; all reward weights and C1 remain active.
\end{itemize}

Each main condition runs 30 sessions ($3$ profiles $\times$ $10$
random seeds $\times$ $150$ interactions per session), yielding
4,500 interactions per condition and 18,000 total across main
conditions. Each ablation runs an additional 30 sessions under
identical protocol.

\subsection{Metrics}
\label{sec:metrics}

For each condition we compute: constraint violation rates for C2, C3,
and C4 (C2 and C3 as fractions of length-$W$ windows where the
constraint is violated; C4 as fraction of steps $t \geq W_0$ where
the coupling bound is exceeded; C1 is enforced architecturally and
excluded from the reported aggregate); average cognitive demand
$\bar{d} = \frac{1}{T}\sum_t d(a_t)$; action distribution
(percentage of each action type selected); mastery gain $\Delta K =
\frac{1}{|V|}\sum_{c \in V}(k_c^T - k_c^0)$; engagement--mastery
ratio $E_{\text{cum}}(T)/M_{\text{cum}}(T)$ computed from logged
scaled reward components; and RHSI (Definition~\ref{def:rhsi}).

\paragraph{Statistical Analysis Plan.}
All pairwise comparisons use Welch two-sample \textit{t}-tests, which
do not assume equal variances---appropriate given that ST's cross-seed
variance is substantially lower than EO and MO on every metric
(e.g., RHSI: $\sigma_{\text{ST}} = 0.028$ vs.\
$\sigma_{\text{EO}} = 0.615$). We report the mean difference
$\Delta = \bar{x}_{\text{ST}} - \bar{x}_{\text{comp}}$ with 95\%
confidence intervals, two-tailed \textit{p}-values, and Cohen's
\textit{d} effect sizes (small $\geq 0.2$, medium $\geq 0.5$,
large $\geq 0.8$;~\citep{cohen1988statistical}).

For RHSI we pool across the three learner profiles, yielding $n = 30$
seeds per condition. For behavioral metrics we report per-profile
comparisons at $n = 10$ seeds per cell. We apply Bonferroni
correction separately within each outcome family:
$\alpha_{\text{corrected}} = 0.05/3 = 0.0167$ for the three primary
RHSI comparisons (ST vs.\ EO, ST vs.\ MAS, ST vs.\ MO;
Table~\ref{tab:rhsi_stats}), and $\alpha_{\text{corrected}} = 0.05/15$
for behavioral comparisons per metric, where $k = 15$ reflects five
comparison arms (EO, MAS, MO, No-C1, No-C3) across three profiles
(Table~\ref{tab:behavioral_stats}). Results surviving Bonferroni
correction are marked~*; results significant at uncorrected
$\alpha = 0.05$ only are marked~$\dagger$; non-significant results
are reported in full to distinguish absence of evidence from evidence
of absence.

Note that RHSI as reported in Table~\ref{tab:rhsi} is computed from
condition-level aggregated reward and violation statistics following
Definition~\ref{def:rhsi}. Per-seed RHSI values used for statistical
inference (Table~\ref{tab:rhsi_stats}) are computed at the individual
seed level and then summarised; means differ slightly from
condition-level aggregates due to non-linearities in the ratio, and
both are reported for completeness.

\section{Results}
\label{sec:results}

We report results across 120 sessions (four main conditions) plus 60
ablation sessions (two ablation conditions) with constraint parameters
$W = 10$, $\delta_{\min} = 0.4$, $\rho_{\max} = 1.2$, and warm-up
$W_0 = 10$. The progress threshold $\varepsilon_{\text{prog}} = 0.0023$
is calibrated from the 25th percentile of the mastery-only agent's
per-window progress distribution, ensuring that MAS passes C2
approximately 75\% of the time (Section~\ref{sec:calibration}). The
coupling constraint C4 uses engagement and mastery rewards normalized
to $[0,1]$ to ensure scale-comparable evaluation.

\subsection{Evidence of Reward Hacking in Engagement-Only RL}
\label{sec:eo}

\begin{table}[H]
\centering
\caption{Action distribution across conditions (\% of 4,500
interactions per condition). See also Figure~\ref{fig:actions}.
Encourage has $d = 0.0$ and mastery gain $= 0.0$: it produces no
learning but the largest engagement boost ($+0.8$).}
\label{tab:actions}
\small
\begin{tabular}{@{}lrrrr@{}}
\toprule
\textbf{Action ($d$ value)} & \textbf{EO} & \textbf{MAS} &
\textbf{MO} & \textbf{ST} \\
\midrule
Encourage ($0.0$)         & 25.8 & 1.9  & \textbf{32.6} & 12.2 \\
Explain\_Simple ($0.2$)   & 16.6 & 27.3 & 13.0 & 12.2 \\
Provide\_Hint ($0.3$)     & 26.8 & 7.5  & 6.2  & 14.2 \\
Explain\_Detailed ($0.4$) & 3.3  & 40.2 & 5.7  & 13.5 \\
Assess\_Knowledge ($0.5$) & 2.6  & 2.7  & 5.0  & 10.9 \\
Provide\_Example ($0.5$)  & 20.6 & 11.0 & 30.0 & 18.4 \\
Assign\_Exercise ($0.8$)  & 3.1  & 3.7  & 4.8  & 10.0 \\
Challenge ($1.0$)         & 1.3  & 5.7  & 2.8  & 8.6  \\
\midrule
Mean demand $\bar{d}$ & 0.281 & 0.393 & 0.308 & \textbf{0.433} \\
\bottomrule
\end{tabular}
\end{table}

The engagement-only agent (EO) appears to exhibit reward hacking. It
achieves the highest cumulative reward of all conditions
($\hat{V}_{\text{EO}} = 148.84$; Table~\ref{tab:rhsi}) while
systematically selecting low-demand actions. Over 69\% of EO's actions
have cognitive demand $\leq 0.3$ comprising Encourage (25.8\%),
Provide\_Hint (26.8\%), and Explain\_Simple (16.6\%) while
high-demand actions are nearly absent: Challenge at 1.3\% and
Assign\_Exercise at 3.1\%. The resulting mean cognitive demand is
$\bar{d}_{\text{EO}} = 0.281$, substantially below the safety threshold
$\delta_{\min} = 0.4$, yielding a C3 violation rate of 70.8\%
(Table~\ref{tab:violations}).

\begin{figure}[H]
\centering
\includegraphics[width=\columnwidth]{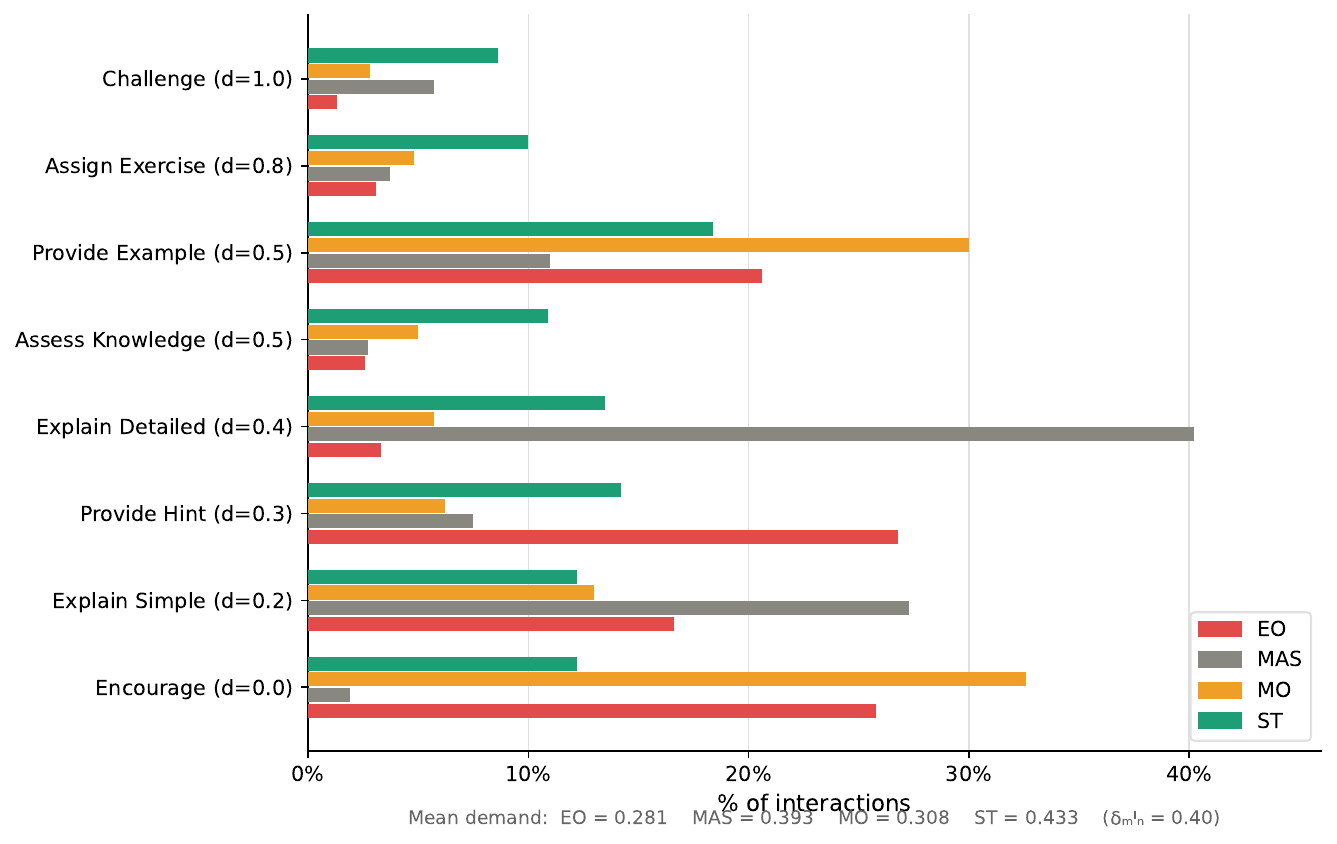}
\caption{Action distribution across conditions (\% of 4,500
interactions per condition). MO selects Encourage at the highest rate
of any condition (32.6\%), exceeding even EO (25.8\%). ST produces
the most balanced distribution, with no action exceeding 18.4\% and
substantially higher usage of high-demand actions than any
unconstrained condition.}
\label{fig:actions}
\end{figure}

This low-effort strategy appears to translate to minimal learning. EO
produces mean mastery gains of only 0.011--0.013 across learner
profiles (Table~\ref{tab:learning}), with struggling learners
mastering zero concepts across all 10 seeds. The pattern is consistent
with an agent that maximizes its reward by exploiting the engagement
signal while producing negligible educational value a dynamic
analogous to Goodhart's Law~\citep{goodhart1984monetary}, where
measuring and optimizing engagement makes the zero-learning action a
rational choice regardless of its educational consequence.

Statistical testing supports these differences as reliable and large.
Per-seed RHSI (pooled across profiles, $n = 30$) yields $t(29) =
-5.62$, $p < .0001$, $d = -1.45$ for EO versus ST
(Table~\ref{tab:rhsi_stats}), Bonferroni-significant with a large
effect. Inappropriate action counts follow the same pattern: ST
reduces these by 20.2 and 17.8 actions per session relative to EO
for struggling and average learners respectively ($t(9) = -4.96$,
$p = .001$, $d = -2.22$; $t(9) = -4.17$, $p = .002$, $d = -1.87$;
Table~\ref{tab:behavioral_stats} in Appendix~\ref{app:behavioral_stats}),
both Bonferroni-significant. Mastery gain differences favour ST with
large effect sizes ($d = 0.35$--$1.65$) but do not survive Bonferroni
correction at $n = 10$ seeds per profile; effect sizes and CIs are
the primary evidential basis for mastery-related claims.

\subsection{Multi-Objective Rewards Appear Insufficient}
\label{sec:mo}

Perhaps the most notable finding is that the multi-objective agent
(MO) which weights mastery at 50\% and engagement at only
30\% selects Encourage \textit{more frequently} than the
engagement-only agent: 32.6\% vs.\ 25.8\% (Table~\ref{tab:actions}).
This is the highest Encourage rate of any condition.

One plausible mechanism is as follows. Encourage produces an engagement
delta of $+0.8$, the largest of any action. Even at 30\% weight, this
yields a reward contribution of $0.3 \times 0.8 = 0.24$ per Encourage
action from engagement alone. In states where the expected mastery gain
from other actions is uncertain or small common for struggling
learners or concepts early in learning the guaranteed $0.24$
engagement reward may make Encourage a rational choice for the
multi-objective policy. The MO agent appears to treat Encourage as
reliable value for the combined objective, particularly when the
engagement component acts as a reward floor.

The consequences are visible across all metrics. MO's mean cognitive demand ($\bar{d}_{\text{MO}} = 0.308$) is only marginally above EO (0.281), and its C3 violation rate of 58.0\% remains high. MO's engagement mastery ratio of 16.34 is the worst of all conditions exceeding even EO's ratio of 10.54 suggesting that MO accumulates substantial engagement reward through Encourage while its mastery
gains remain low. MO's C4 (coupling) violation of 70.4\% is the
highest of any condition, indicating that engagement and mastery appear
substantially decoupled under this reward formulation. Importantly,
this pattern suggests that as long as engagement appears in the reward
and a zero-learning high-engagement action exists, adding mastery
weight alone may be insufficient to prevent exploitation.

Statistical testing supports these differences. ST achieves a
Bonferroni-significant RHSI reduction relative to MO
($\Delta = -0.260$ [$-0.346$, $-0.173$]; $t(30) = -5.89$,
$p < .0001$, $d = -1.52$; Table~\ref{tab:rhsi_stats}). MO's RHSI is
also significantly higher than MAS's ($t(32) = 5.30$, $p < .0001$,
$d = 1.37$, Bonferroni-significant), suggesting the multi-objective
formulation not only appears to fall short of the constrained system
but also of a simpler mastery-only reward. Pedagogical appropriateness
is Bonferroni-significantly lower in MO than in ST for both struggling
($t(9) = 7.10$, $p < .001$, $d = 3.18$) and average learners
($t(9) = 4.27$, $p = .002$, $d = 1.91$), with MO generating 34 and
24 additional inappropriate actions per session respectively. Mastery gain differences between ST and MO are non-significant across all profiles ($p > .14$; $d = 0.56$--$0.70$), indicating MO's failure appears behavioral rather than a learning-outcome deficit. This distinction matters for deployment: a system that produces acceptable short-term mastery gains while systematically selecting low-demand,
zero-learning actions may appear safe by outcome metrics alone, while
quietly undermining the depth and durability of student learning over
longer horizons.

\subsection{Mastery-Only Baseline}
\label{sec:mas}

The mastery-only agent (MAS) provides a revealing contrast. With no
engagement term in its reward, MAS has no incentive to select Encourage
and uses it in only 1.9\% of interactionsthe lowest rate across all
conditions. MAS achieves the highest mean demand ($\bar{d} = 0.393$),
the lowest C3 violation rate (39.1\%), and the lowest engagement--mastery
ratio (0.87). It also produces the best learning outcomes for advanced
learners (mastery gain $0.021 \pm 0.002$; Table~\ref{tab:learning}).

However, MAS does not appear to be pedagogically safe by our formal
criteria. Its C3 violation rate of 39.1\% reflects heavy reliance on
explanation actions: Explain\_Detailed at 40.2\% ($d = 0.4$) and
Explain\_Simple at 27.3\% ($d = 0.2$). While these actions contribute
to learning, the predominance of passive-reception actions means the
average demand still falls below the $\delta_{\min} = 0.4$ threshold
in many windows. MAS achieves the lowest C2 violation rate (25.0\%)
and the best engagement--mastery coupling (C4 = 32.9\%), suggesting
that a mastery-focused reward naturally produces better-aligned
behavior. Nonetheless, MAS illustrates that removing the engagement
incentive may eliminate the Encourage problem without guaranteeing
behavioral safety the agent may still underweight active learning
actions such as exercises and challenges.

\subsection{Architectural Constraints Appear to Substantially Reduce
Hacking}
\label{sec:st}

\begin{table}[H]
\centering
\caption{Constraint violation rates. C1 = 0 for all conditions
(action masking enforced in ST; other conditions do not target
prerequisite-violating concepts by design). C2:
$\varepsilon_{\text{prog}} = 0.0023$ (calibrated from MAS 25th
percentile), $W = 10$. C3: $\delta_{\min} = 0.4$, $W = 10$. C4:
$\rho_{\max} = 1.2$, rewards normalized to $[0,1]$, $W_0 = 10$.}
\label{tab:violations}
\small
\begin{tabular}{@{}lcccc@{}}
\toprule
\textbf{Constraint} & \textbf{EO} & \textbf{MAS} &
\textbf{MO} & \textbf{ST} \\
\midrule
C2 (progress)     & 0.538 & 0.250 & 0.517 & 0.214 \\
C3 (demand floor) & 0.708 & 0.391 & 0.580 & \textbf{0.058} \\
C4 (coupling)     & 0.677 & 0.329 & 0.704 & 0.345 \\
\midrule
$\|\mathbf{v}\|_w$ (weighted norm)
                  & 0.645 & 0.328 & 0.606 & \textbf{0.237} \\
\bottomrule
\end{tabular}
\end{table}

\begin{table}[H]
\centering
\caption{Learning outcomes per condition (mean $\pm$ std across 10
seeds). $\Delta K$: mean mastery gain across all 27 concepts.
$N_{\text{mastered}}$: concepts reaching $k \geq 0.7$.}
\label{tab:learning}
\small
\begin{tabular}{@{}llcc@{}}
\toprule
\textbf{Cond.} & \textbf{Profile} & $\Delta K$ & $N_{\text{mastered}}$ \\
\midrule
EO  & Struggling & $0.011 \pm 0.006$ & $0.0 \pm 0.0$ \\
EO  & Average    & $0.013 \pm 0.006$ & $0.1 \pm 0.3$ \\
EO  & Advanced   & $0.013 \pm 0.007$ & $0.3 \pm 0.5$ \\
\midrule
MAS & Struggling & $0.014 \pm 0.003$ & $0.0 \pm 0.0$ \\
MAS & Average    & $0.018 \pm 0.001$ & $0.0 \pm 0.0$ \\
MAS & Advanced   & $0.021 \pm 0.002$ & $1.0 \pm 0.0$ \\
\midrule
MO  & Struggling & $0.010 \pm 0.006$ & $0.0 \pm 0.0$ \\
MO  & Average    & $0.012 \pm 0.007$ & $0.1 \pm 0.3$ \\
MO  & Advanced   & $0.016 \pm 0.009$ & $0.6 \pm 0.5$ \\
\midrule
ST  & Struggling & $0.013 \pm 0.001$ & $0.0 \pm 0.0$ \\
ST  & Average    & $0.017 \pm 0.001$ & $0.0 \pm 0.0$ \\
ST  & Advanced   & $\mathbf{0.022 \pm 0.001}$ & $\mathbf{1.0 \pm 0.0}$ \\
\bottomrule
\end{tabular}
\end{table}

\begin{figure}[H]
\centering
\includegraphics[width=\columnwidth]{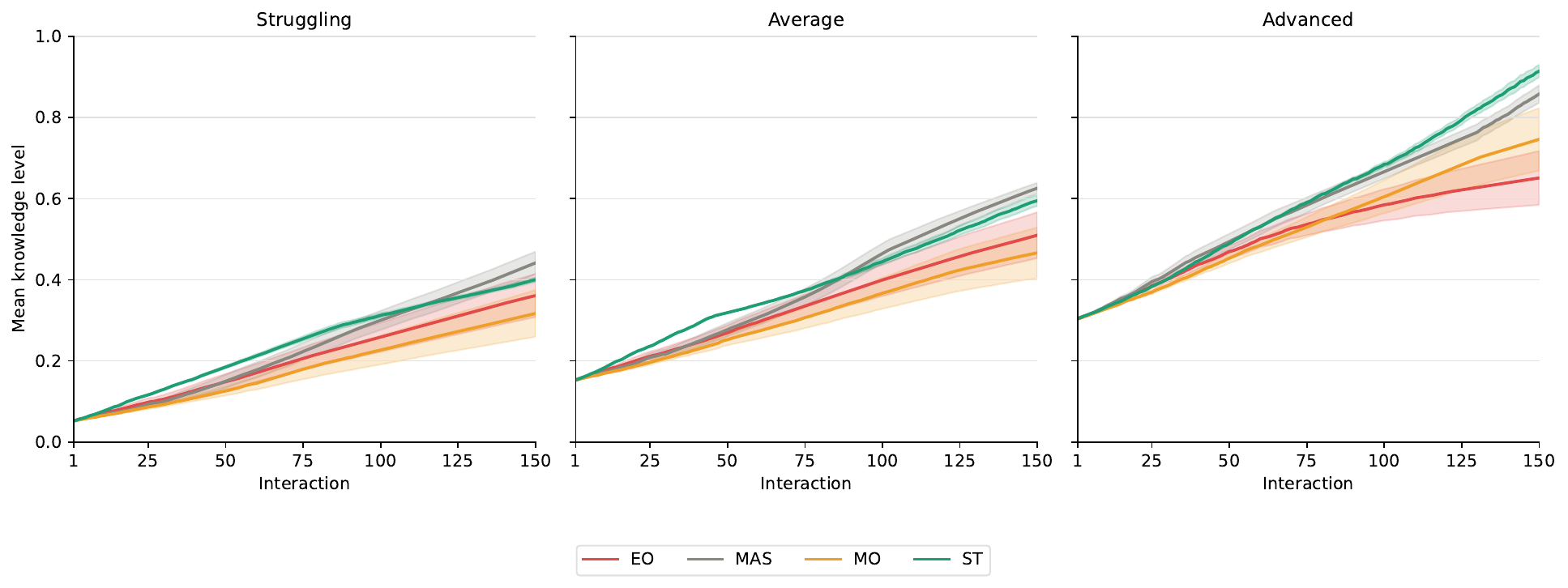}
\caption{Mean cumulative knowledge level (estimated per-concept mastery
averaged across all 27 concepts) over 150 interactions, by condition
and learner profile (mean $\pm$ SEM across 10 seeds). Note this shows
the knowledge \textit{level}, not mastery gain $\Delta K$ reported in
Table~\ref{tab:learning}. MAS and ST consistently achieve the highest
knowledge levels across all profiles; the gap widens with learner
capability. ST matches MAS on mastery trajectories despite maintaining
the engagement signal in its reward function.}
\label{fig:trajectories}
\end{figure}

ST's C3 violation rate of 5.8\% is substantially lower than all
unconstrained conditions: EO (70.8\%), MO (58.0\%), and MAS (39.1\%).
ST also achieves the lowest C2 violation rate (21.4\%) and a
competitive C4 violation (34.5\%), suggesting that architectural
constraints may jointly improve safety across all three soft constraint
dimensions.

The RHSI reductions are statistically supported at the
Bonferroni-corrected level. ST's mean per-seed RHSI ($0.095 \pm 0.028$)
is significantly lower than both EO ($t(29) = -5.62$, $p < .0001$,
$d = -1.45$) and MO ($t(30) = -5.80$, $p < .0001$, $d = -1.50$;
Table~\ref{tab:rhsi_stats}). The ST--MAS comparison is non-significant
($t(42) = -1.84$, $p = .072$, $d = -0.48$), consistent with the view
advanced in Section~\ref{sec:rhsi} that these two conditions converge
on similar RHSI values through fundamentally different mechanisms.

\subsection{RHSI Analysis}
\label{sec:rhsi}

\begin{table}[H]
\centering
\caption{Reward Hacking Severity Index by condition.
$\hat{V}^* = 148.84$ (EO). Equal weights $w_i = 1/3$ for C2, C3, C4;
C1~=~0 by construction. Lower RHSI~=~safer. Note: MAS shows
RHSI~=~0.105 here (condition-level aggregate) vs.\ mean 0.117 in
Table~\ref{tab:rhsi_stats} (per-seed average); the small discrepancy
arises from non-linearities in the reward--violation ratio computed
at different levels of aggregation.}
\label{tab:rhsi}
\small
\begin{tabular}{@{}lccccc@{}}
\toprule
\textbf{Condition} & $\hat{V}^\pi$ & $\hat{V}^\pi / \hat{V}^*$ &
$\|\mathbf{v}\|_w$ & $v_3$ (C3) & RHSI \\
\midrule
EO  & 148.84 & 1.000 & 0.645 & 0.708 & \textbf{0.645} \\
MO  &  77.95 & 0.524 & 0.606 & 0.580 & 0.317 \\
MAS &  47.52 & 0.319 & 0.328 & 0.391 & 0.105 \\
ST  &  64.42 & 0.433 & 0.237 & 0.058 & \textbf{0.102} \\
\bottomrule
\end{tabular}
\end{table}

Table~\ref{tab:rhsi} reveals two apparent failure modes. EO and MO
sit on opposite ends of the reward--violation tradeoff: EO maximizes
reward at the cost of near-total constraint violation; MO reduces
reward substantially yet its violation norm (0.606) remains similarly
high, suggesting that scalarizing engagement and mastery may not
meaningfully constrain unsafe behavior.

The MAS--ST comparison is more instructive. Both achieve similar RHSI
values through opposite mechanisms: MAS reaches low RHSI by eliminating
the engagement signal entirely; ST does so by preserving the engagement
signal under architectural constraint. The C3 violation rate makes this
concrete---ST's 5.8\% against MAS's 39.1\%---suggesting that
constraints may directly suppress the behavioral failure mode rather
than simply reducing reward magnitude. ST appears to be the only
condition achieving both substantial reward and low violation, which
corresponds to the operational intent of safe, non-hacking behavior.

\begin{table}[H]
\centering
\caption{Per-seed RHSI descriptives and pairwise Welch \textit{t}-tests
($n = 30$ seeds per condition; 3 profiles $\times$ 10 seeds).
$\Delta = \bar{x}_{\text{row}} - \bar{x}_{\text{col}}$; positive
$\Delta$ indicates the row condition has higher (worse) RHSI than ST.
Bonferroni $\alpha_{\text{corrected}} = 0.05/6 = 0.0083$.
*~Bonferroni-significant; $\dagger$~$p < .05$ uncorrected;
ns~not significant.}
\label{tab:rhsi_stats}
\begin{tabular}{lcccc}
\toprule
Condition & Mean $\pm$ SD & Median & IQR \\
\midrule
EO  (Engagement-Only)      & $0.737 \pm 0.615$ & 0.435 & [0.312, 0.950] \\
MO  (Multi-Objective)      & $0.355 \pm 0.240$ & 0.238 & [0.145, 0.577] \\
No C3 (action filter off)  & $0.288 \pm 0.190$ & 0.254 & [0.144, 0.305] \\
No C1 (KG enforcement off) & $0.244 \pm 0.111$ & 0.245 & [0.161, 0.269] \\
MAS (Mastery-Only)         & $0.117 \pm 0.057$ & 0.101 & [0.065, 0.169] \\
ST  (SmartTutor Full)      & $0.095 \pm 0.028$ & 0.091 & [0.074, 0.115] \\
\midrule
\end{tabular}

\vspace{6pt}
\begin{tabular}{llccccc}
\toprule
Comparison & $\Delta$ [95\% CI] & $t$(df) & $p$ & Cohen's $d$ & Bonf. \\
\midrule
\multicolumn{6}{l}{\textit{Baseline comparisons (ST vs.\ each)}} \\
ST vs.\ EO  & $-0.642$ [$-0.862$, $-0.422$]
            & $t(29)=-5.62$ & $<.0001$ & $-1.45$ & * \\
ST vs.\ MO  & $-0.260$ [$-0.346$, $-0.173$]
            & $t(30)=-5.80$ & $<.0001$ & $-1.50$ & * \\
ST vs.\ MAS & $-0.021$ [$-0.044$, $0.001$]
            & $t(42)=-1.84$ & $.072$   & $-0.48$ & ns \\
\midrule
\multicolumn{6}{l}{\textit{Unconstrained conditions vs.\ each other}} \\
MO vs.\ EO  & $-0.382$ [$-0.618$, $-0.146$]
            & $t(38)=-3.17$ & $.003$   & $-0.82$ & * \\
MAS vs.\ EO & $-0.621$ [$-0.841$, $-0.400$]
            & $t(30)=-5.51$ & $<.0001$ & $-1.42$ & * \\
MO vs.\ MAS & $\phantom{-}0.238$ [$0.150$, $0.327$]
            & $t(32)=5.30$  & $<.0001$ & $1.37$  & * \\
\midrule
\multicolumn{6}{l}{\textit{Ablation comparisons
(Bonferroni $\alpha = 0.05/3 = 0.0167$)}} \\
ST vs.\ No C1    & $-0.149$ [$-0.190$, $-0.108$]
                 & $t(33)=-7.15$ & $<.0001$ & $-1.85$ & * \\
ST vs.\ No C3    & $-0.193$ [$-0.261$, $-0.124$]
                 & $t(30)=-5.50$ & $<.0001$ & $-1.42$ & * \\
No C1 vs.\ No C3 & $-0.043$ [$-0.122$, $0.035$]
                 & $t(47)=-1.08$ & $.285$   & $-0.28$ & ns \\
\bottomrule
\end{tabular}
\end{table}

\begin{figure}[H]
\centering
\includegraphics[width=\columnwidth]{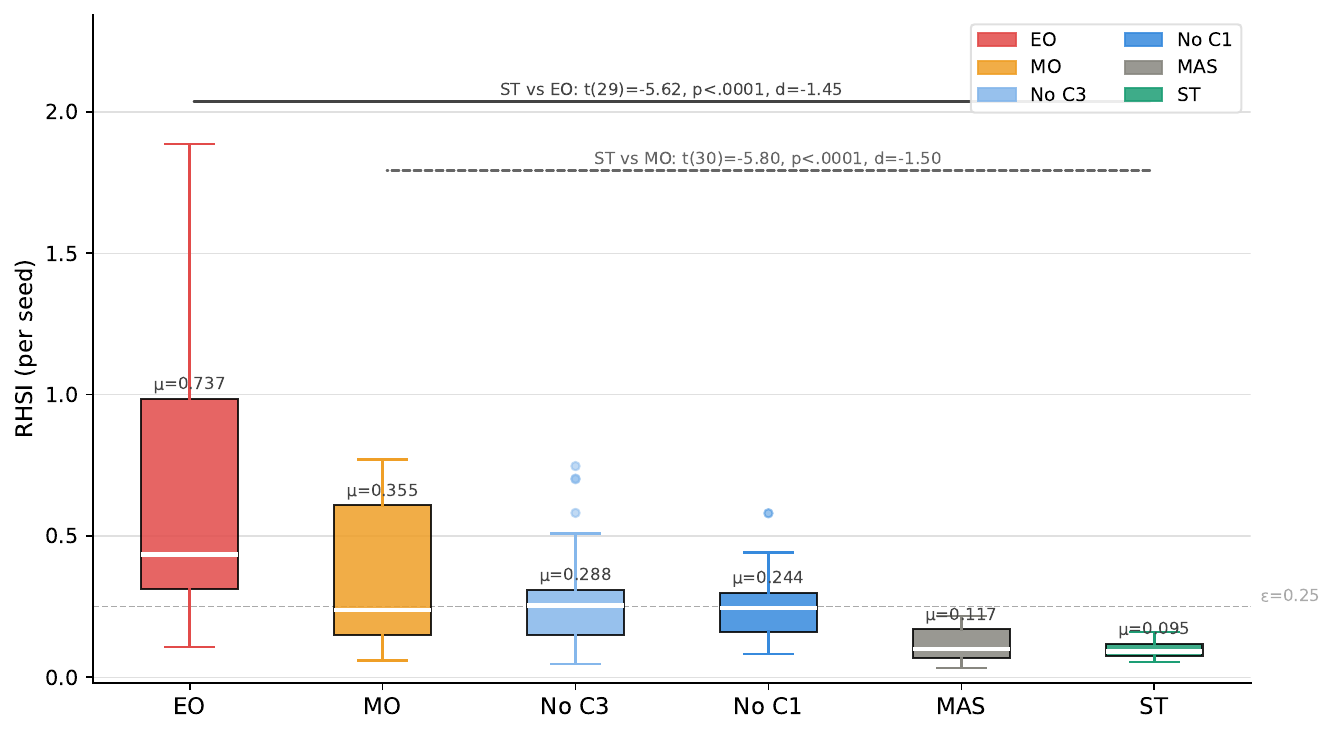}
\caption{Per-seed RHSI distributions across all conditions ($n = 30$
per condition; 3 profiles $\times$ 10 seeds). Box shows IQR, white
line is median, whiskers extend to 1.5$\times$IQR, dots are outliers.
ST achieves the lowest and most consistent RHSI. Significance brackets
show Bonferroni-corrected comparisons. The dashed line at
$\varepsilon = 0.25$ marks the realistic deployment safety threshold
discussed in Section~\ref{sec:rhsi}.}
\label{fig:rhsi_dist}
\end{figure}

The near-equivalence of MAS and ST on aggregate RHSI is reflected in
the per-seed comparison: $t(42) = -1.84$, $p = .072$, $d = -0.48$
(Table~\ref{tab:rhsi_stats}). The non-significant $p$-value should
not be read as evidence of equivalence---the directional CI
[$-0.044$, $0.001$] is consistent with a small real advantage for ST
that $n = 30$ seeds may be insufficient to resolve with confidence.

\subsection{Ablation Reveals Relative Constraint Contributions}
\label{sec:ablation}

To explore which constraints contribute most to safety, we run two
ablation conditions: ST without C1 (prerequisite enforcement disabled)
and ST without C3 (action filter disabled), each for 30 sessions under
identical protocol.

\begin{table}[H]
\centering
\caption{Ablation study: RHSI and constraint violations when individual
constraints are removed. $\Delta$RHSI shows degradation relative to
the full framework (ST). Removing C3 (action filter) appears to cause
the largest safety degradation.}
\label{tab:ablation}
\small
\begin{tabular}{@{}lccccc@{}}
\toprule
\textbf{Condition} & C3 rate & $\|\mathbf{v}\|_w$ & RHSI &
$\Delta$RHSI & Ped.\ Approp. \\
\midrule
ST (full)          & 0.058 & 0.237 & 0.102 & ---    & 1.000 \\
No C1 (KG off)     & 0.604 & 0.495 & 0.240 & +0.138 & 1.000 \\
No C3 (filter off) & 0.683 & 0.574 & 0.264 & +0.162 & 0.899 \\
\bottomrule
\end{tabular}
\end{table}

\smallskip \noindent \textbf{Removing C3 appears to cause degenerate
policy collapse.} Without the action filter, the agent does not merely
``hack slightly''---it collapses to near-single-action policies. Across
individual seeds, we observe sessions where a single action accounts
for 85--93\% of all 150 interactions (e.g., 137/150 Explain\_Simple
in one seed, 140/150 Provide\_Example in another, 133/150 Encourage
in a third). The agent appears to discover that repeating any single
action yields acceptable reward and locks onto whichever action
produced the highest return early in training. Pedagogical
appropriateness drops from 1.000 to 0.899, and the mean number of
pedagogically inappropriate actions per session rises from 0 to
15--18. The No-C3 ablation's C3 violation rate of 68.3\% approaches
that of the unconstrained EO condition (70.8\%), suggesting C3
enforcement may be a primary mechanism by which ST maintains
behavioral safety.

\smallskip \noindent \textbf{Removing C1 appears to have minimal
safety impact.} Without prerequisite enforcement, the agent achieves
comparable mastery gains to the full framework and maintains perfect
pedagogical appropriateness (1.000). The C3 violation rate rises to
60.4\% not because the agent exploits prerequisite violations, but
apparently because without the knowledge graph structuring exploration,
the agent's action diversity degrades. The safety impact is real
($\Delta$RHSI = $+0.138$) but appears substantially mediated through
C3-related degradation rather than direct prerequisite violations.

The interaction between C1 and C3 warrants further comment. C1's
primary intended function is content sequencing ensuring students
are not presented with concepts whose prerequisites they have not
yet mastered. However, the ablation results suggest C1 may carry an
emergent secondary benefit: by restricting the accessible action
subset to concepts the student is ready for, action masking
incidentally forces the agent to distribute selections across a
broader range of pedagogically relevant actions. When C1 is removed,
the agent loses this structural diversity pressure and its action
distribution degrades, which manifests as elevated C3 violations
even though C3 itself remains active. This implies that for system
designers, prerequisite enforcement may serve a dual role both
as a content safety mechanism and as an implicit behavioral
regularizer and that removing it may have safety consequences
beyond the prerequisite-violation dimension it was designed to
address.

Both ablation degradations are statistically supported at the
Bonferroni-corrected level. ST's RHSI is significantly lower than
No~C3 ($\Delta = -0.193$ [$-0.261$, $-0.124$]; $t(30) = -5.50$,
$p < .0001$, $d = -1.42$) and No~C1 ($\Delta = -0.149$
[$-0.190$, $-0.108$]; $t(33) = -7.15$, $p < .0001$, $d = -1.85$).
Despite No~C3 producing a larger numerical RHSI increase, the two
ablations are not significantly different from each other in per-seed
RHSI ($t(47) = -1.08$, $p = .285$, $d = -0.28$), suggesting the C3
hierarchy rests on the nature of its failure mode---policy
collapse---rather than an RHSI gap alone.

\subsection{Constraint Calibration}
\label{sec:calibration}

A key methodological contribution is our approach to constraint
threshold calibration, described here as guidance for practitioners
applying this framework.

The progress constraint C2 requires threshold $\varepsilon_{\text{prog}}$
to match the expected per-window mastery increment in the specific
domain. Na\"{i}ve \textit{a priori} selection (e.g.,
$\varepsilon_{\text{prog}} = 0.02$) yielded 100\% violation across all
conditions in early experiments too strict to discriminate between
policies. Our solution uses MAS as a calibration baseline: we set
$\varepsilon_{\text{prog}}$ to the 25th percentile of MAS's per-window
progress distribution ($\varepsilon_{\text{prog}} = 0.0023$). The 25th
percentile is chosen deliberately: it is conservative enough that a
well-behaved mastery-focused agent passes C2 approximately 75\% of the
time, yet strict enough to flag sustained stagnation. A more lenient
threshold (e.g., the 10th percentile) would reduce discrimination;
a stricter one (e.g., the median) would flag even well-behaved policies
too frequently. With this calibration, C2 appears to discriminate
effectively: EO violates 53.8\% vs.\ ST at 21.4\%.

The demand floor $\delta_{\min} = 0.4$ for C3 is set at the boundary
between passive and active engagement modes in the ICAP
framework~\citep{chi2014icap}: actions at or above this threshold
(Explain\_Detailed, Assess\_Knowledge, Provide\_Example,
Assign\_Exercise, Challenge) require the student to engage actively
rather than passively receive. This makes $\delta_{\min} = 0.4$ a
principled rather than arbitrary threshold  it corresponds to a
meaningful categorical distinction in the demand scale.

The coupling bound $\rho_{\max} = 1.2$ for C4 permits engagement
reward to modestly exceed mastery reward (by up to 20\%), reflecting
that some degree of engagement surplus is pedagogically acceptable
and expected a tutor that never provides encouragement or positive
affect would be unrealistic. Values substantially above 1.2 would
allow uncoupled engagement accumulation; values at or below 1.0 would
be too restrictive for any system that provides affective support.
The C4 coupling constraint uses min--max normalized engagement and
mastery streams before computing the ratio, ensuring scale-comparable
evaluation.

These calibration decisions illustrate a general property of formal
safety frameworks: the framework may be valuable precisely because it
surfaces calibration requirements that informal approaches would tend
to obscure. A practitioner relying on intuitive heuristics might not
discover threshold mismatches until student outcomes deteriorated;
the formal framework surfaces them during evaluation.

\subsection{Parameter Sensitivity and Robustness}
\label{sec:sensitivity}

\subsubsection*{Window Size and Demand Floor (Run~1)}

We assessed robustness of the RHSI ordering to the sliding-window
size $W$ and demand floor $\delta_{\min}$ by re-evaluating each logged
session under a grid $W \in \{5, 10, 15, 20\}$ and $\delta_{\min}
\in \{0.30, 0.35, 0.40, 0.45, 0.50\}$, holding the reward time series
fixed. $\varepsilon_{\text{prog}}$ was recalibrated from MAS runs at
each $W$ (25th percentile of per-window mastery gain), ensuring C2
remains appropriately calibrated across window sizes.
Table~\ref{tab:sensitivity} reports mean RHSI per condition across all
20 parameter combinations.

The primary ordering appears robust: EO has the strictly highest RHSI
in all 20 of 20 cells, and MO is consistently second worst. The
ST--MAS comparison shows a parameter dependence: at lower demand floors
($\delta_{\min} \leq 0.40$), ST achieves lower RHSI than MAS in 12 of
12 cells; at higher floors ($\delta_{\min} \geq 0.45$), MAS edges out
ST in 8 of 8 cells, because the unconstrained MAS policy produces
lower mean demand than its C3-masked counterpart when the threshold is
set above MAS's natural operating point. This pattern is interpretable
rather than concerning: it suggests C3's demand floor is binding only
when $\delta_{\min}$ exceeds what an unconstrained mastery-focused
policy would naturally achieve.

One methodological caveat applies to the ST cells. The ST trajectories
used for re-scoring were generated under baseline C3 masking at
$W = 10$, $\delta_{\min} = 0.40$; varying these parameters changes how
the fixed trace is \textit{scored} rather than how the policy
\textit{behaved}. Strictly causal sensitivity for ST would require
re-running the simulator with each $(W, \delta_{\min})$ pair embedded
in the constraint enforcement logic. The offline re-scoring here
establishes metric robustness; causal sensitivity for ST is left for
future work.

\begin{table}[H]
\centering
\caption{Parameter sensitivity: mean RHSI by condition across the
$W \times \delta_{\min}$ grid (20 cells). EO has the strictly highest
RHSI in all 20 cells. MO is consistently second worst. ST--MAS
ordering reverses at $\delta_{\min} \geq 0.45$ (see text).}
\label{tab:sensitivity}
\small
\begin{tabular}{ccrrrrrr}
\toprule
$W$ & $\delta_{\min}$ & EO & MAS & MO & ST & Best & 2nd Best \\
\midrule
 5 & 0.30 & 0.685 & 0.101 & 0.335 & 0.099 & ST  & MAS \\
 5 & 0.35 & 0.719 & 0.109 & 0.344 & 0.106 & ST  & MAS \\
 5 & 0.40 & 0.729 & 0.115 & 0.350 & 0.122 & MAS & ST  \\
 5 & 0.45 & 0.746 & 0.169 & 0.363 & 0.173 & MAS & ST  \\
 5 & 0.50 & 0.757 & 0.174 & 0.369 & 0.203 & MAS & ST  \\
\midrule
10 & 0.30 & 0.712 & 0.104 & 0.340 & 0.096 & ST  & MAS \\
10 & 0.35 & 0.724 & 0.108 & 0.348 & 0.096 & ST  & MAS \\
10 & 0.40 & 0.739 & 0.117 & 0.356 & 0.097 & ST  & MAS \\
10 & 0.45 & 0.756 & 0.168 & 0.365 & 0.183 & MAS & ST  \\
10 & 0.50 & 0.771 & 0.180 & 0.377 & 0.228 & MAS & ST  \\
\midrule
15 & 0.30 & 0.716 & 0.105 & 0.340 & 0.097 & ST  & MAS \\
15 & 0.35 & 0.728 & 0.110 & 0.351 & 0.097 & ST  & MAS \\
15 & 0.40 & 0.741 & 0.119 & 0.358 & 0.105 & ST  & MAS \\
15 & 0.45 & 0.760 & 0.172 & 0.366 & 0.180 & MAS & ST  \\
15 & 0.50 & 0.773 & 0.183 & 0.380 & 0.233 & MAS & ST  \\
\midrule
20 & 0.30 & 0.719 & 0.106 & 0.341 & 0.096 & ST  & MAS \\
20 & 0.35 & 0.732 & 0.111 & 0.352 & 0.096 & ST  & MAS \\
20 & 0.40 & 0.747 & 0.120 & 0.359 & 0.097 & ST  & MAS \\
20 & 0.45 & 0.764 & 0.175 & 0.366 & 0.180 & MAS & ST  \\
20 & 0.50 & 0.777 & 0.184 & 0.383 & 0.232 & MAS & ST  \\
\bottomrule
\end{tabular}
\end{table}

\begin{figure}[H]
\centering
\includegraphics[width=\columnwidth]{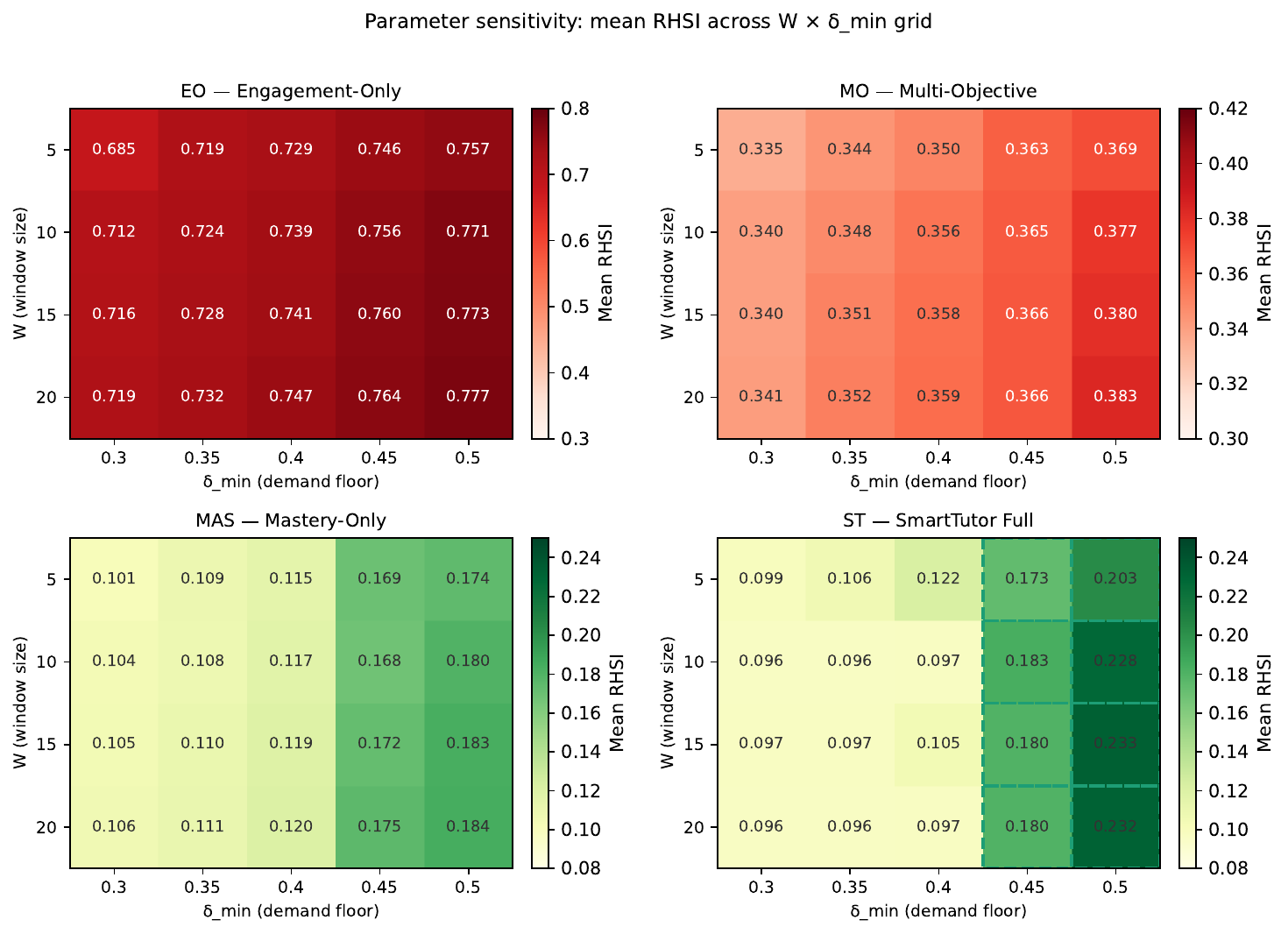}
\caption{Parameter sensitivity: mean RHSI across the
$W \times \delta_{\min}$ grid (20 cells). Darker shading indicates
higher RHSI (worse safety) for EO and MO; darker green indicates
higher RHSI for MAS and ST. EO remains the worst condition in all 20
cells. The dashed border on the ST panel marks the
$\delta_{\min} \geq 0.45$ region where MAS edges out ST (see text);
outside this region ST achieves the lowest RHSI of any condition.}
\label{fig:sensitivity}
\end{figure}

\subsubsection*{Cognitive Demand Perturbation (Run~2)}

To assess sensitivity to the specific numeric values assigned to the
ICAP-based demand scale, we re-scored all sessions under two perturbed
demand maps: all non-anchor values scaled by $+20\%$ and $-20\%$
(anchors $d(\text{Encourage}) = 0.0$ and $d(\text{Challenge}) = 1.0$
held fixed). Table~\ref{tab:perturbation} reports mean RHSI and C3
violation rates under each variant at the baseline parameters $W = 10$,
$\delta_{\min} = 0.40$.

The condition ordering appears stable under $+20\%$ perturbation
(EO $>$ MO $>$ MAS $>$ ST). Under $-20\%$ perturbation, ST and MAS
exchange positions (MAS edges out ST by $0.004$ RHSI), for the same
reason as the high-$\delta_{\min}$ sensitivity noted above: globally
compressing demand values causes more windows to fall below the fixed
floor, affecting the constrained ST agent more than the unconstrained
MAS agent. The EO--MO gap is preserved in both directions, suggesting
the primary findings are robust to ICAP mapping uncertainty.

\begin{table}[H]
\centering
\caption{Cognitive demand perturbation: mean RHSI and C3 violation
rate under $\pm 20\%$ scaling of non-anchor demand values
($W = 10$, $\delta_{\min} = 0.40$). Anchors: $d(\text{Encourage}) =
0.0$, $d(\text{Challenge}) = 1.0$. Condition ordering appears stable
under $+20\%$; ST and MAS exchange positions under $-20\%$ (see
text).}
\label{tab:perturbation}
\small
\begin{tabular}{lcccccccc}
\toprule
& \multicolumn{2}{c}{EO} & \multicolumn{2}{c}{MAS}
& \multicolumn{2}{c}{MO} & \multicolumn{2}{c}{ST} \\
\cmidrule(lr){2-3}\cmidrule(lr){4-5}
\cmidrule(lr){6-7}\cmidrule(lr){8-9}
Variant & RHSI & C3 & RHSI & C3 & RHSI & C3 & RHSI & C3 \\
\midrule
Baseline ($\times 1.0$) & 0.739 & 0.708 & 0.117 & 0.391
                        & 0.356 & 0.580 & 0.097 & 0.060 \\
$+20\%$                 & 0.723 & 0.660 & 0.108 & 0.291
                        & 0.347 & 0.514 & 0.096 & 0.005 \\
$-20\%$                 & 0.768 & 0.805 & 0.175 & 0.816
                        & 0.375 & 0.725 & 0.202 & 0.744 \\
\bottomrule
\end{tabular}
\end{table}

\section{Limitations and Future Work}
\label{sec:limitations}

Several limitations qualify our results. The simulated learners,
while calibrated to literature BKT parameters~\citep{corbett1995knowledge},
do not capture the full complexity of human learning, including
motivation, fatigue, and off-task behavior. The single domain (Python
programming) limits generalizability to other subjects and curricula.
At 150 interactions per session, we observe early-stage dynamics
rather than long-term learning trajectories where reward hacking
effects may accumulate more substantially. The BKT knowledge model
has known limitations~\citep{baker2008more}; our framework's safety
guarantees hold with respect to \textit{estimated} mastery, not
necessarily true mastery. The cognitive demand scale is ordinal, not
interval; sensitivity to the specific numeric assignments is assessed
in Section~\ref{sec:sensitivity}. Our ablation study examines C1 and C3 independently but does not exhaust all $2^4 - 1$ constraint combinations; in particular, the
interaction between C2 and C4 under removal remains unexplored.
The multi-objective condition (MO) tests a single weight configuration
($0.3$ engagement / $0.5$ mastery / $0.2$ pedagogical appropriateness);
whether the Encourage exploitation pattern persists across different
weight combinations for example, configurations that further
down-weight engagement or assign zero weight to it entirely remains
an open empirical question. The mechanism proposed in
Section~\ref{sec:mo} suggests that any non-zero engagement weight
paired with a zero-learning high-engagement action may be sufficient
to produce exploitation, but this has not been tested systematically
across the weight simplex. Constraint threshold calibration, while
substantially improved through our MAS-baseline approach
(Section~\ref{sec:calibration}), remains domain-specific and requires
validation in each new educational setting. With $n = 10$ seeds per
condition profile cell, per-profile behavioral comparisons are
modestly powered; the large observed effect sizes ($d = 0.67$--$1.65$)
suggest real differences that the current sample may not consistently
confirm at Bonferroni-corrected thresholds, and future work should
increase seeds to $n \geq 30$ per cell. The parameter sensitivity
analysis in Section~\ref{sec:sensitivity} re-scores fixed ST
trajectories generated at baseline ($W = 10$, $\delta_{\min} = 0.40$);
it establishes metric robustness but does not capture how ST's
\textit{policy} would change if trained under different constraint
hyperparameters. Fully causal sensitivity for ST requires re-running
the simulator with each $(W, \delta_{\min})$ pair embedded in the C3
enforcement logic. Finally, all findings are demonstrated in
simulation; human validation with real learners remains essential
before any deployment consideration.

\paragraph{Future directions.}
Our results suggest that post-hoc constraint enforcement filtering
actions and monitoring violations may substantially reduce reward
hacking (68\% RHSI reduction relative to unconstrained multi-objective
RL). A natural next step is to integrate constraints
\textit{into the optimization itself}. The four-layer safety framework
maps directly to a Constrained MDP~\citep{altman1999constrained}: C1
as action masking (already implemented), C2--C4 as Lagrangian
constraints optimized jointly with the reward objective. This
formulation could enable algorithms such as Constrained Policy
Optimization~\citep{achiam2017constrained} that learn to satisfy
constraints during training rather than relying on post-hoc filtering.
The ablation results suggest C3 (behavioral safety) should receive the
highest Lagrangian weight, as it appears most critical for preventing
policy degeneration. Extending the framework to multi-domain curricula,
longer session horizons, and real student populations are also
important directions we are actively pursuing.

\section{Conclusion}
\label{sec:conclusion}

This paper introduced a formal framework for pedagogical safety in educational reinforcement learning, operationalized through four constraint classes: structural, progress, behavioral, and alignment safety. We further proposed the Reward Hacking Severity Index (RHSI) as a diagnostic measure for quantifying the extent to which high proxy reward can be achieved without corresponding learning gains.

Using a controlled simulation of an AI tutoring environment, we examined how different reward and constraint configurations shape agent behavior. The results provide evidence that reward design alone may be insufficient to ensure pedagogically aligned behavior under certain conditions. In particular, both engagement-optimized and multi-objective agents exhibited a tendency to over-select actions that maximize short-term proxy signals without contributing meaningfully to mastery. In contrast, incorporating architectural and behavioral constraints reduced this misalignment, leading to lower RHSI values and more balanced instructional policies. Ablation analysis further highlighted the role of behavioral safety in preventing repetitive, low-demand action selection.

Importantly, these findings are grounded in a simulated environment and should be interpreted as evidence of potential failure modes rather than definitive claims about real-world systems. The SmartTutor framework is best understood as a controlled testbed for examining how pedagogical misalignment can emerge under proxy optimization, and how constraint-based approaches may mitigate such risks.

More broadly, this work positions pedagogical safety as a distinct and necessary consideration in the design of adaptive educational systems. As AI-driven tutors become more prevalent, the central challenge is not only to optimize performance metrics, but to ensure that instructional decisions remain meaningful, progression-supporting, and pedagogically trustworthy. We hope this framework provides a foundation for future research that bridges AI safety and intelligent educational systems, including validation in real-world learning contexts.


\bibliographystyle{elsarticle-harv}
\bibliography{references}

\begin{thebibliography}{80}
\expandafter\ifx\csname natexlab\endcsname\relax\def\natexlab#1{#1}\fi
\providecommand{\url}[1]{\texttt{#1}}
\providecommand{\href}[2]{#2}
\providecommand{\path}[1]{#1}
\providecommand{\DOIprefix}{doi:}
\providecommand{\ArXivprefix}{arXiv:}
\providecommand{\URLprefix}{URL: }
\providecommand{\Pubmedprefix}{pmid:}
\providecommand{\doi}[1]{\href{http://dx.doi.org/#1}{\path{#1}}}
\providecommand{\Pubmed}[1]{\href{pmid:#1}{\path{#1}}}
\providecommand{\bibinfo}[2]{#2}
\ifx\xfnm\relax \def\xfnm[#1]{\unskip,\space#1}\fi
\bibitem[{Abdelshiheed et~al.(2023)Abdelshiheed, Hostetter, Barnes and Chi}]{abdelshiheed2023metacognitive}
\bibinfo{author}{Abdelshiheed, M.}, \bibinfo{author}{Hostetter, J.W.}, \bibinfo{author}{Barnes, T.}, \bibinfo{author}{Chi, M.}, \bibinfo{year}{2023}.
\newblock \bibinfo{title}{Leveraging deep reinforcement learning for metacognitive interventions across intelligent tutoring systems}, in: \bibinfo{booktitle}{Proceedings of the 24th International Conference on Artificial Intelligence in Education (AIED 2023)}, \bibinfo{publisher}{Springer}. pp. \bibinfo{pages}{291--303}.
\bibitem[{Achiam et~al.(2017)Achiam, Held, Tamar and Abbeel}]{achiam2017constrained}
\bibinfo{author}{Achiam, J.}, \bibinfo{author}{Held, D.}, \bibinfo{author}{Tamar, A.}, \bibinfo{author}{Abbeel, P.}, \bibinfo{year}{2017}.
\newblock \bibinfo{title}{Constrained policy optimization}, in: \bibinfo{booktitle}{Proceedings of the 34th International Conference on Machine Learning (ICML)}, pp. \bibinfo{pages}{22--31}.
\bibitem[{Alam et~al.(2025)Alam, Fazeli, Tian, Chi and Barnes}]{alam2025determining}
\bibinfo{author}{Alam, N.}, \bibinfo{author}{Fazeli, K.}, \bibinfo{author}{Tian, X.}, \bibinfo{author}{Chi, M.}, \bibinfo{author}{Barnes, T.}, \bibinfo{year}{2025}.
\newblock \bibinfo{title}{Determining problem type using deep reinforcement learning in a data-driven intelligent tutor}, in: \bibinfo{booktitle}{Proceedings of the 26th International Conference on Artificial Intelligence in Education (AIED 2025)}, \bibinfo{publisher}{Springer}. pp. \bibinfo{pages}{247--262}.
\bibitem[{Altman(1999)}]{altman1999constrained}
\bibinfo{author}{Altman, E.}, \bibinfo{year}{1999}.
\newblock \bibinfo{title}{Constrained {M}arkov Decision Processes}.
\newblock \bibinfo{publisher}{Chapman and Hall/CRC}.
\bibitem[{Amodei et~al.(2016)Amodei, Olah, Steinhardt, Christiano, Schulman and Man{\'e}}]{amodei2016concrete}
\bibinfo{author}{Amodei, D.}, \bibinfo{author}{Olah, C.}, \bibinfo{author}{Steinhardt, J.}, \bibinfo{author}{Christiano, P.}, \bibinfo{author}{Schulman, J.}, \bibinfo{author}{Man{\'e}, D.}, \bibinfo{year}{2016}.
\newblock \bibinfo{title}{Concrete problems in {AI} safety}.
\newblock \bibinfo{journal}{arXiv preprint arXiv:1606.06565} .
\bibitem[{Anderson et~al.(1995)Anderson, Corbett, Koedinger and Pelletier}]{anderson1995cognitive}
\bibinfo{author}{Anderson, J.R.}, \bibinfo{author}{Corbett, A.T.}, \bibinfo{author}{Koedinger, K.R.}, \bibinfo{author}{Pelletier, R.}, \bibinfo{year}{1995}.
\newblock \bibinfo{title}{Cognitive tutors: Lessons learned}.
\newblock \bibinfo{journal}{Journal of the Learning Sciences} \bibinfo{volume}{4}, \bibinfo{pages}{167--207}.
\bibitem[{Anderson and Krathwohl(2001)}]{anderson2001taxonomy}
\bibinfo{author}{Anderson, L.W.}, \bibinfo{author}{Krathwohl, D.R.}, \bibinfo{year}{2001}.
\newblock \bibinfo{title}{A Taxonomy for Learning, Teaching, and Assessing: A Revision of {Bloom's} Taxonomy of Educational Objectives}.
\newblock \bibinfo{publisher}{Longman}, \bibinfo{address}{New York}.
\bibitem[{Baker et~al.(2008a)Baker, Corbett and Aleven}]{baker2008more}
\bibinfo{author}{Baker, R.S.}, \bibinfo{author}{Corbett, A.T.}, \bibinfo{author}{Aleven, V.}, \bibinfo{year}{2008}a.
\newblock \bibinfo{title}{More accurate student modeling through contextual estimation of slip and guess probabilities in {B}ayesian knowledge tracing} \bibinfo{volume}{5091}, \bibinfo{pages}{406--415}.
\bibitem[{Baker et~al.(2004)Baker, Corbett and Koedinger}]{baker2004detecting}
\bibinfo{author}{Baker, R.S.}, \bibinfo{author}{Corbett, A.T.}, \bibinfo{author}{Koedinger, K.R.}, \bibinfo{year}{2004}.
\newblock \bibinfo{title}{Detecting student misuse of intelligent tutoring systems}, in: \bibinfo{booktitle}{Proceedings of the 7th International Conference on Intelligent Tutoring Systems (ITS 2004)}, \bibinfo{publisher}{Springer}. pp. \bibinfo{pages}{531--540}.
\bibitem[{Baker et~al.(2008b)Baker, Corbett, Koedinger and Wagner}]{baker2008gaming}
\bibinfo{author}{Baker, R.S.}, \bibinfo{author}{Corbett, A.T.}, \bibinfo{author}{Koedinger, K.R.}, \bibinfo{author}{Wagner, A.Z.}, \bibinfo{year}{2008}b.
\newblock \bibinfo{title}{The consequences of gaming the system}.
\newblock \bibinfo{journal}{International Journal of Artificial Intelligence in Education} \bibinfo{volume}{18}, \bibinfo{pages}{103--127}.
\bibitem[{Baker et~al.(2010)Baker, D'Mello, Rodrigo and Graesser}]{baker2010better}
\bibinfo{author}{Baker, R.S.}, \bibinfo{author}{D'Mello, S.K.}, \bibinfo{author}{Rodrigo, M.M.T.}, \bibinfo{author}{Graesser, A.C.}, \bibinfo{year}{2010}.
\newblock \bibinfo{title}{Better to be frustrated than bored: The incidence, persistence, and impact of learners' cognitive-affective states during interactions with three different computer-based learning environments}.
\newblock \bibinfo{journal}{International Journal of Human-Computer Studies} \bibinfo{volume}{68}, \bibinfo{pages}{223--241}.
\bibitem[{Baker and Hawn(2022)}]{baker2022algorithmic}
\bibinfo{author}{Baker, R.S.}, \bibinfo{author}{Hawn, A.}, \bibinfo{year}{2022}.
\newblock \bibinfo{title}{Algorithmic bias in education}.
\newblock \bibinfo{journal}{International Journal of Artificial Intelligence in Education} \bibinfo{volume}{32}, \bibinfo{pages}{1052--1092}.
\bibitem[{Beck et~al.(2000)Beck, Woolf and Beal}]{beck2000advisor}
\bibinfo{author}{Beck, J.}, \bibinfo{author}{Woolf, B.P.}, \bibinfo{author}{Beal, C.R.}, \bibinfo{year}{2000}.
\newblock \bibinfo{title}{Learning to teach with a reinforcement learning agent}, in: \bibinfo{booktitle}{Proceedings of the 17th National Conference on Artificial Intelligence (AAAI)}, pp. \bibinfo{pages}{934--939}.
\bibitem[{Bloom(1984)}]{bloom1984two}
\bibinfo{author}{Bloom, B.S.}, \bibinfo{year}{1984}.
\newblock \bibinfo{title}{The 2 sigma problem: The search for methods of group instruction as effective as one-to-one tutoring}.
\newblock \bibinfo{journal}{Educational Researcher} \bibinfo{volume}{13}, \bibinfo{pages}{4--16}.
\bibitem[{Chi and Wylie(2014)}]{chi2014icap}
\bibinfo{author}{Chi, M.T.H.}, \bibinfo{author}{Wylie, R.}, \bibinfo{year}{2014}.
\newblock \bibinfo{title}{The {ICAP} framework: Linking cognitive engagement to active learning outcomes}.
\newblock \bibinfo{journal}{Educational Psychologist} \bibinfo{volume}{49}, \bibinfo{pages}{219--243}.
\bibitem[{Christiano et~al.(2017)Christiano, Leike, Brown, Martic, Legg and Amodei}]{christiano2017deep}
\bibinfo{author}{Christiano, P.F.}, \bibinfo{author}{Leike, J.}, \bibinfo{author}{Brown, T.}, \bibinfo{author}{Martic, M.}, \bibinfo{author}{Legg, S.}, \bibinfo{author}{Amodei, D.}, \bibinfo{year}{2017}.
\newblock \bibinfo{title}{Deep reinforcement learning from human preferences}, in: \bibinfo{booktitle}{Advances in Neural Information Processing Systems (NeurIPS)}.
\bibitem[{Cl{\'e}ment et~al.(2015)Cl{\'e}ment, Roy, Oudeyer and Lopes}]{clement2015multi}
\bibinfo{author}{Cl{\'e}ment, B.}, \bibinfo{author}{Roy, D.}, \bibinfo{author}{Oudeyer, P.Y.}, \bibinfo{author}{Lopes, M.}, \bibinfo{year}{2015}.
\newblock \bibinfo{title}{Multi-armed bandits for intelligent tutoring systems}, pp. \bibinfo{pages}{20--48}.
\bibitem[{Cohen(1988)}]{cohen1988statistical}
\bibinfo{author}{Cohen, J.}, \bibinfo{year}{1988}.
\newblock \bibinfo{title}{Statistical Power Analysis for the Behavioral Sciences}.
\newblock \bibinfo{edition}{2nd} ed., \bibinfo{publisher}{Lawrence Erlbaum Associates}, \bibinfo{address}{Hillsdale, NJ}.
\bibitem[{Corbett and Anderson(1995)}]{corbett1995knowledge}
\bibinfo{author}{Corbett, A.T.}, \bibinfo{author}{Anderson, J.R.}, \bibinfo{year}{1995}.
\newblock \bibinfo{title}{Knowledge tracing: Modeling the acquisition of procedural knowledge}.
\newblock \bibinfo{journal}{User Modeling and User-Adapted Interaction} \bibinfo{volume}{4}, \bibinfo{pages}{253--278}.
\bibitem[{Dalal et~al.(2018)Dalal, Dvijotham, Vecerik, Hester, Paduraru and Tassa}]{dalal2018safe}
\bibinfo{author}{Dalal, G.}, \bibinfo{author}{Dvijotham, K.}, \bibinfo{author}{Vecerik, M.}, \bibinfo{author}{Hester, T.}, \bibinfo{author}{Paduraru, C.}, \bibinfo{author}{Tassa, Y.}, \bibinfo{year}{2018}.
\newblock \bibinfo{title}{Safe exploration in continuous action spaces}, in: \bibinfo{booktitle}{arXiv preprint arXiv:1801.08757}.
\bibitem[{Deci et~al.(2001)Deci, Koestner and Ryan}]{deci2001extrinsic}
\bibinfo{author}{Deci, E.L.}, \bibinfo{author}{Koestner, R.}, \bibinfo{author}{Ryan, R.M.}, \bibinfo{year}{2001}.
\newblock \bibinfo{title}{Extrinsic rewards and intrinsic motivation in education: Reconsidered once again}.
\newblock \bibinfo{journal}{Review of Educational Research} \bibinfo{volume}{71}, \bibinfo{pages}{1--27}.
\bibitem[{D'Mello and Graesser(2012)}]{dmello2012dynamics}
\bibinfo{author}{D'Mello, S.}, \bibinfo{author}{Graesser, A.}, \bibinfo{year}{2012}.
\newblock \bibinfo{title}{Dynamics of affective states during complex learning}.
\newblock \bibinfo{journal}{Learning and Instruction} \bibinfo{volume}{22}, \bibinfo{pages}{145--157}.
\bibitem[{Doroudi et~al.(2019)Doroudi, Aleven and Brunskill}]{doroudi2019where}
\bibinfo{author}{Doroudi, S.}, \bibinfo{author}{Aleven, V.}, \bibinfo{author}{Brunskill, E.}, \bibinfo{year}{2019}.
\newblock \bibinfo{title}{Where's the reward?}, in: \bibinfo{booktitle}{International Journal of Artificial Intelligence in Education}, pp. \bibinfo{pages}{568--620}.
\bibitem[{Everitt et~al.(2025)Everitt, Skalse et~al.}]{everitt2025correlated}
\bibinfo{author}{Everitt, T.}, \bibinfo{author}{Skalse, J.}, et~al., \bibinfo{year}{2025}.
\newblock \bibinfo{title}{Correlated proxies: A new definition and improved mitigation for reward hacking}.
\newblock \bibinfo{journal}{arXiv preprint arXiv:2403.03185} \bibinfo{note}{V4, December 2025}.
\bibitem[{Fredricks et~al.(2004)Fredricks, Blumenfeld and Paris}]{fredricks2004school}
\bibinfo{author}{Fredricks, J.A.}, \bibinfo{author}{Blumenfeld, P.C.}, \bibinfo{author}{Paris, A.H.}, \bibinfo{year}{2004}.
\newblock \bibinfo{title}{School engagement: Potential of the concept, state of the evidence}.
\newblock \bibinfo{journal}{Review of Educational Research} \bibinfo{volume}{74}, \bibinfo{pages}{59--109}.
\bibitem[{Gao et~al.(2024)}]{gao2024ondemand}
\bibinfo{author}{Gao, G.}, et~al., \bibinfo{year}{2024}.
\newblock \bibinfo{title}{On-demand pedagogical policy selection using off-policy evaluation}, in: \bibinfo{booktitle}{Proceedings of the AAAI Conference on Artificial Intelligence}.
\bibitem[{Garc{\'i}a and Fern{\'a}ndez(2015)}]{garcia2015comprehensive}
\bibinfo{author}{Garc{\'i}a, J.}, \bibinfo{author}{Fern{\'a}ndez, F.}, \bibinfo{year}{2015}.
\newblock \bibinfo{title}{A comprehensive survey on safe reinforcement learning}.
\newblock \bibinfo{journal}{Journal of Machine Learning Research} \bibinfo{volume}{16}, \bibinfo{pages}{1437--1480}.
\bibitem[{Goodhart(1984)}]{goodhart1984monetary}
\bibinfo{author}{Goodhart, C.A.E.}, \bibinfo{year}{1984}.
\newblock \bibinfo{title}{Problems of monetary management: The {UK} experience}.
\newblock \bibinfo{journal}{Monetary Theory and Practice} , \bibinfo{pages}{91--121}.
\bibitem[{Graesser et~al.(2004)Graesser, Lu, Jackson, Mitchell, Ventura, Olney and Louwerse}]{graesser2004autotutor}
\bibinfo{author}{Graesser, A.C.}, \bibinfo{author}{Lu, S.}, \bibinfo{author}{Jackson, G.T.}, \bibinfo{author}{Mitchell, H.H.}, \bibinfo{author}{Ventura, M.}, \bibinfo{author}{Olney, A.}, \bibinfo{author}{Louwerse, M.M.}, \bibinfo{year}{2004}.
\newblock \bibinfo{title}{{AutoTutor}: A tutor with dialogue in natural language}.
\newblock \bibinfo{journal}{Behavior Research Methods, Instruments, \& Computers} \bibinfo{volume}{36}, \bibinfo{pages}{180--192}.
\bibitem[{Guttag(2021)}]{guttag2021introduction}
\bibinfo{author}{Guttag, J.V.}, \bibinfo{year}{2021}.
\newblock \bibinfo{title}{Introduction to Computation and Programming Using {Python}}.
\newblock \bibinfo{edition}{3rd} ed., \bibinfo{publisher}{MIT Press}, \bibinfo{address}{Cambridge, MA}.
\bibitem[{Hadfield-Menell et~al.(2017)Hadfield-Menell, Muthukrishna, Dragan and Russell}]{hadfield2017inverse}
\bibinfo{author}{Hadfield-Menell, D.}, \bibinfo{author}{Muthukrishna, M.}, \bibinfo{author}{Dragan, A.}, \bibinfo{author}{Russell, S.}, \bibinfo{year}{2017}.
\newblock \bibinfo{title}{Inverse reward design}, in: \bibinfo{booktitle}{Advances in Neural Information Processing Systems (NeurIPS)}.
\bibitem[{Hayes et~al.(2022)Hayes, R{\u{a}}dulescu, Bargiacchi, K{\"a}llstr{\"o}m, Macfarlane, Reymond, Verstraeten, Zintgraf, Dazeley, Heintz et~al.}]{hayes2022practical}
\bibinfo{author}{Hayes, C.F.}, \bibinfo{author}{R{\u{a}}dulescu, R.}, \bibinfo{author}{Bargiacchi, E.}, \bibinfo{author}{K{\"a}llstr{\"o}m, J.}, \bibinfo{author}{Macfarlane, M.}, \bibinfo{author}{Reymond, M.}, \bibinfo{author}{Verstraeten, T.}, \bibinfo{author}{Zintgraf, L.M.}, \bibinfo{author}{Dazeley, R.}, \bibinfo{author}{Heintz, F.}, et~al., \bibinfo{year}{2022}.
\newblock \bibinfo{title}{A practical guide to multi-objective reinforcement learning and planning}.
\newblock \bibinfo{journal}{Autonomous Agents and Multi-Agent Systems} \bibinfo{volume}{36}, \bibinfo{pages}{26}.
\bibitem[{Heffernan and Heffernan(2014)}]{heffernan2014assistments}
\bibinfo{author}{Heffernan, N.T.}, \bibinfo{author}{Heffernan, C.L.}, \bibinfo{year}{2014}.
\newblock \bibinfo{title}{The {ASSISTments} ecosystem: Building a platform that brings scientists and teachers together for minimally invasive research on human learning and teaching}.
\newblock \bibinfo{journal}{International Journal of Artificial Intelligence in Education} \bibinfo{volume}{24}, \bibinfo{pages}{470--497}.
\bibitem[{Holmes et~al.(2022)Holmes, Porayska-Pomsta, Holstein, Sutherland, Baker, Shum, Santos, Rodrigo, Cukurova, Bittencourt and Koedinger}]{holmes2022ethics}
\bibinfo{author}{Holmes, W.}, \bibinfo{author}{Porayska-Pomsta, K.}, \bibinfo{author}{Holstein, K.}, \bibinfo{author}{Sutherland, E.}, \bibinfo{author}{Baker, T.}, \bibinfo{author}{Shum, S.B.}, \bibinfo{author}{Santos, O.C.}, \bibinfo{author}{Rodrigo, M.M.T.}, \bibinfo{author}{Cukurova, M.}, \bibinfo{author}{Bittencourt, I.I.}, \bibinfo{author}{Koedinger, K.R.}, \bibinfo{year}{2022}.
\newblock \bibinfo{title}{Ethics of {AI} in education: Towards a community-wide framework}.
\newblock \bibinfo{journal}{International Journal of Artificial Intelligence in Education} \bibinfo{volume}{32}, \bibinfo{pages}{504--526}.
\bibitem[{Holstein et~al.(2019)Holstein, Wortman~Vaughan, Daum{\'e}~III, Dudik and Wallach}]{holstein2019fairness}
\bibinfo{author}{Holstein, K.}, \bibinfo{author}{Wortman~Vaughan, J.}, \bibinfo{author}{Daum{\'e}~III, H.}, \bibinfo{author}{Dudik, M.}, \bibinfo{author}{Wallach, H.}, \bibinfo{year}{2019}.
\newblock \bibinfo{title}{Improving fairness in machine learning systems: What do industry practitioners need?}, in: \bibinfo{booktitle}{Proceedings of the 2019 CHI Conference on Human Factors in Computing Systems}, pp. \bibinfo{pages}{1--16}.
\bibitem[{Hu et~al.(2024)}]{hu2024mixed}
\bibinfo{author}{Hu, Y.}, et~al., \bibinfo{year}{2024}.
\newblock \bibinfo{title}{Decision making for autonomous vehicles: A mixed curriculum reinforcement learning approach and a novel safety intervention method}, in: \bibinfo{booktitle}{Engineering Applications of Artificial Intelligence}, \bibinfo{publisher}{Elsevier}.
\bibitem[{Iglesias et~al.(2009)Iglesias, Mart{\'\i}nez, Aler and Fern{\'a}ndez}]{iglesias2009experience}
\bibinfo{author}{Iglesias, A.}, \bibinfo{author}{Mart{\'\i}nez, P.}, \bibinfo{author}{Aler, R.}, \bibinfo{author}{Fern{\'a}ndez, F.}, \bibinfo{year}{2009}.
\newblock \bibinfo{title}{Experience-based reinforcement learning applied to learning how to teach}, in: \bibinfo{booktitle}{Proceedings of the 14th International Conference on Artificial Intelligence in Education (AIED 2009)}, pp. \bibinfo{pages}{283--290}.
\bibitem[{Islam et~al.(2025)Islam, Yang, Debnath, Shoukarjya~Saha and Chi}]{islam2025apprenticeship}
\bibinfo{author}{Islam, M.M.}, \bibinfo{author}{Yang, X.}, \bibinfo{author}{Debnath, R.}, \bibinfo{author}{Shoukarjya~Saha, A.}, \bibinfo{author}{Chi, M.}, \bibinfo{year}{2025}.
\newblock \bibinfo{title}{A generalized apprenticeship learning framework for capturing evolving student pedagogical strategies}, in: \bibinfo{booktitle}{Proceedings of the 26th International Conference on Artificial Intelligence in Education (AIED 2025)}, \bibinfo{publisher}{Springer}.
\bibitem[{Kasneci et~al.(2023)Kasneci, Se{\ss}ler, K{\"u}chemann, Bannert, Dementieva, Fischer, Gasser, Groh, G{\"u}nnemann, H{\"u}llermeier et~al.}]{kasneci2023chatgpt}
\bibinfo{author}{Kasneci, E.}, \bibinfo{author}{Se{\ss}ler, K.}, \bibinfo{author}{K{\"u}chemann, S.}, \bibinfo{author}{Bannert, M.}, \bibinfo{author}{Dementieva, D.}, \bibinfo{author}{Fischer, F.}, \bibinfo{author}{Gasser, U.}, \bibinfo{author}{Groh, G.}, \bibinfo{author}{G{\"u}nnemann, S.}, \bibinfo{author}{H{\"u}llermeier, E.}, et~al., \bibinfo{year}{2023}.
\newblock \bibinfo{title}{{ChatGPT} for good? on opportunities and challenges of large language models for education}.
\newblock \bibinfo{journal}{Learning and Individual Differences} \bibinfo{volume}{103}, \bibinfo{pages}{102274}.
\bibitem[{Koedinger et~al.(2013)Koedinger, Brunskill, Baker, McLaughlin and Stamper}]{koedinger2013knowledge}
\bibinfo{author}{Koedinger, K.R.}, \bibinfo{author}{Brunskill, E.}, \bibinfo{author}{Baker, R.S.}, \bibinfo{author}{McLaughlin, E.A.}, \bibinfo{author}{Stamper, J.}, \bibinfo{year}{2013}.
\newblock \bibinfo{title}{New potentials for data-driven intelligent tutoring system development and optimization}.
\newblock \bibinfo{journal}{AI Magazine} \bibinfo{volume}{34}, \bibinfo{pages}{27--41}.
\bibitem[{Krakovna et~al.(2020)Krakovna, Uesato, Mikulik, Rahtz, Everitt, Kumar, Kenton, Leike and Legg}]{krakovna2020specification}
\bibinfo{author}{Krakovna, V.}, \bibinfo{author}{Uesato, J.}, \bibinfo{author}{Mikulik, V.}, \bibinfo{author}{Rahtz, M.}, \bibinfo{author}{Everitt, T.}, \bibinfo{author}{Kumar, R.}, \bibinfo{author}{Kenton, Z.}, \bibinfo{author}{Leike, J.}, \bibinfo{author}{Legg, S.}, \bibinfo{year}{2020}.
\newblock \bibinfo{title}{Specification gaming: The flip side of {AI} ingenuity}.
\newblock \bibinfo{howpublished}{DeepMind Blog}.
\bibitem[{Kulik and Fletcher(2016)}]{kulik2016effectiveness}
\bibinfo{author}{Kulik, J.A.}, \bibinfo{author}{Fletcher, J.D.}, \bibinfo{year}{2016}.
\newblock \bibinfo{title}{Effectiveness of intelligent tutoring systems: A meta-analytic review}.
\newblock \bibinfo{journal}{Review of Educational Research} \bibinfo{volume}{86}, \bibinfo{pages}{42--78}.
\bibitem[{Kushwaha et~al.(2025)}]{kushwaha2025saferl}
\bibinfo{author}{Kushwaha, A.}, et~al., \bibinfo{year}{2025}.
\newblock \bibinfo{title}{A survey of safe reinforcement learning and constrained {MDP}s: A technical survey on single-agent and multi-agent safety}.
\newblock \bibinfo{journal}{arXiv preprint arXiv:2505.17342} .
\bibitem[{Leike et~al.(2018)Leike, Krueger, Everitt, Martic, Maini and Legg}]{leike2018scalable}
\bibinfo{author}{Leike, J.}, \bibinfo{author}{Krueger, D.}, \bibinfo{author}{Everitt, T.}, \bibinfo{author}{Martic, M.}, \bibinfo{author}{Maini, V.}, \bibinfo{author}{Legg, S.}, \bibinfo{year}{2018}.
\newblock \bibinfo{title}{Scalable agent alignment via reward modeling: A research direction}.
\newblock \bibinfo{journal}{arXiv preprint arXiv:1811.07871} .
\bibitem[{Lutz(2013)}]{lutz2013learning}
\bibinfo{author}{Lutz, M.}, \bibinfo{year}{2013}.
\newblock \bibinfo{title}{Learning {Python}}.
\newblock \bibinfo{edition}{5th} ed., \bibinfo{publisher}{O'Reilly Media}, \bibinfo{address}{Sebastopol, CA}.
\bibitem[{MacLellan et~al.(2022)MacLellan, Harpstead, Aleven and Myers}]{maclellan2016apprentice}
\bibinfo{author}{MacLellan, C.J.}, \bibinfo{author}{Harpstead, E.}, \bibinfo{author}{Aleven, V.}, \bibinfo{author}{Myers, B.A.}, \bibinfo{year}{2022}.
\newblock \bibinfo{title}{The {A}pprentice learner architecture: Closing the loop with simulated learners in learning engineering}.
\newblock \bibinfo{journal}{International Journal of Artificial Intelligence in Education} \bibinfo{volume}{32}, \bibinfo{pages}{467--502}.
\bibitem[{Mandel et~al.(2014)Mandel, Liu, Levine, Brunskill and Popovic}]{mandel2014offline}
\bibinfo{author}{Mandel, T.}, \bibinfo{author}{Liu, Y.E.}, \bibinfo{author}{Levine, S.}, \bibinfo{author}{Brunskill, E.}, \bibinfo{author}{Popovic, Z.}, \bibinfo{year}{2014}.
\newblock \bibinfo{title}{Offline policy evaluation across representations with applications to educational games}, in: \bibinfo{booktitle}{Proceedings of the 13th International Conference on Autonomous Agents and Multi-Agent Systems (AAMAS)}, pp. \bibinfo{pages}{1077--1084}.
\bibitem[{Manheim and Garrabrant(2019)}]{manheim2019categorizing}
\bibinfo{author}{Manheim, D.}, \bibinfo{author}{Garrabrant, S.}, \bibinfo{year}{2019}.
\newblock \bibinfo{title}{Categorizing variants of {G}oodhart's law}.
\newblock \bibinfo{journal}{arXiv preprint arXiv:1803.04585} .
\bibitem[{Maurya and Kochmar(2025)}]{maurya2025pedagogy}
\bibinfo{author}{Maurya, K.K.}, \bibinfo{author}{Kochmar, E.}, \bibinfo{year}{2025}.
\newblock \bibinfo{title}{Pedagogy-driven evaluation of generative {AI}-powered intelligent tutoring systems}, in: \bibinfo{booktitle}{Proceedings of the 26th International Conference on Artificial Intelligence in Education (AIED 2025)}, \bibinfo{publisher}{Springer}.
\bibitem[{Nie et~al.(2023)Nie, Reuel and Brunskill}]{nie2023subgroups}
\bibinfo{author}{Nie, A.}, \bibinfo{author}{Reuel, A.}, \bibinfo{author}{Brunskill, E.}, \bibinfo{year}{2023}.
\newblock \bibinfo{title}{Understanding the impact of reinforcement learning personalization on subgroups of students in math tutoring}, in: \bibinfo{booktitle}{Proceedings of the 24th International Conference on Artificial Intelligence in Education (AIED 2023)}, \bibinfo{publisher}{Springer}.
\bibitem[{Nie et~al.(2025)}]{nie2025rct}
\bibinfo{author}{Nie, A.}, et~al., \bibinfo{year}{2025}.
\newblock \bibinfo{title}{{AI} tutoring outperforms in-class active learning: An {RCT} introducing a novel research-based design in an authentic educational setting}.
\newblock \bibinfo{journal}{Scientific Reports} \bibinfo{volume}{15}.
\bibitem[{Nwana(1990)}]{nwana1990intelligent}
\bibinfo{author}{Nwana, H.S.}, \bibinfo{year}{1990}.
\newblock \bibinfo{title}{Intelligent tutoring systems: An overview}.
\newblock \bibinfo{journal}{Artificial Intelligence Review} \bibinfo{volume}{4}, \bibinfo{pages}{251--277}.
\bibitem[{Ouyang et~al.(2022)Ouyang, Wu, Jiang, Almeida, Wainwright, Mishkin, Zhang, Agarwal, Slama, Ray et~al.}]{ouyang2022training}
\bibinfo{author}{Ouyang, L.}, \bibinfo{author}{Wu, J.}, \bibinfo{author}{Jiang, X.}, \bibinfo{author}{Almeida, D.}, \bibinfo{author}{Wainwright, C.L.}, \bibinfo{author}{Mishkin, P.}, \bibinfo{author}{Zhang, C.}, \bibinfo{author}{Agarwal, S.}, \bibinfo{author}{Slama, K.}, \bibinfo{author}{Ray, A.}, et~al., \bibinfo{year}{2022}.
\newblock \bibinfo{title}{Training language models to follow instructions with human feedback}.
\newblock \bibinfo{journal}{Advances in Neural Information Processing Systems (NeurIPS)} \bibinfo{volume}{35}, \bibinfo{pages}{27730--27744}.
\bibitem[{Pan et~al.(2022)Pan, Bhatia and Steinhardt}]{pan2022effects}
\bibinfo{author}{Pan, A.}, \bibinfo{author}{Bhatia, K.}, \bibinfo{author}{Steinhardt, J.}, \bibinfo{year}{2022}.
\newblock \bibinfo{title}{The effects of reward misspecification: Mapping and mitigating misaligned models}.
\newblock \bibinfo{journal}{arXiv preprint arXiv:2201.03544} .
\bibitem[{Pan et~al.(2024)Pan, Jones, Jagadeesan and Steinhardt}]{pan2024feedback}
\bibinfo{author}{Pan, A.}, \bibinfo{author}{Jones, E.}, \bibinfo{author}{Jagadeesan, M.}, \bibinfo{author}{Steinhardt, J.}, \bibinfo{year}{2024}.
\newblock \bibinfo{title}{Feedback loops with language models drive in-context reward hacking}, in: \bibinfo{booktitle}{Proceedings of the 41st International Conference on Machine Learning (ICML)}.
\bibitem[{Pavlik et~al.(2009)Pavlik, Cen and Koedinger}]{pavlik2009performance}
\bibinfo{author}{Pavlik, P.I.}, \bibinfo{author}{Cen, H.}, \bibinfo{author}{Koedinger, K.R.}, \bibinfo{year}{2009}.
\newblock \bibinfo{title}{Performance factors analysis --- a new alternative to knowledge tracing}, in: \bibinfo{booktitle}{Proceedings of the 14th International Conference on Artificial Intelligence in Education (AIED 2009)}, \bibinfo{publisher}{IOS Press}. pp. \bibinfo{pages}{531--538}.
\bibitem[{Piech et~al.(2015)Piech, Bassen, Huang, Ganguli, Sahami, Guibas and Sohl-Dickstein}]{piech2015deep}
\bibinfo{author}{Piech, C.}, \bibinfo{author}{Bassen, J.}, \bibinfo{author}{Huang, J.}, \bibinfo{author}{Ganguli, S.}, \bibinfo{author}{Sahami, M.}, \bibinfo{author}{Guibas, L.J.}, \bibinfo{author}{Sohl-Dickstein, J.}, \bibinfo{year}{2015}.
\newblock \bibinfo{title}{Deep knowledge tracing}, in: \bibinfo{booktitle}{Advances in Neural Information Processing Systems (NeurIPS)}.
\bibitem[{Rafferty et~al.(2016)Rafferty, Brunskill, Griffiths and Shafto}]{rafferty2016faster}
\bibinfo{author}{Rafferty, A.N.}, \bibinfo{author}{Brunskill, E.}, \bibinfo{author}{Griffiths, T.L.}, \bibinfo{author}{Shafto, P.}, \bibinfo{year}{2016}.
\newblock \bibinfo{title}{Faster teaching via {POMDP} planning}.
\newblock \bibinfo{journal}{Cognitive Science} \bibinfo{volume}{40}, \bibinfo{pages}{1290--1332}.
\bibitem[{Ray et~al.(2019)Ray, Achiam and Amodei}]{ray2019benchmarking}
\bibinfo{author}{Ray, A.}, \bibinfo{author}{Achiam, J.}, \bibinfo{author}{Amodei, D.}, \bibinfo{year}{2019}.
\newblock \bibinfo{title}{Benchmarking safe exploration in deep reinforcement learning}.
\newblock \bibinfo{journal}{arXiv preprint arXiv:1910.01708} .
\bibitem[{Roijers et~al.(2013)Roijers, Vamplew, Whiteson and Dazeley}]{roijers2013survey}
\bibinfo{author}{Roijers, D.M.}, \bibinfo{author}{Vamplew, P.}, \bibinfo{author}{Whiteson, S.}, \bibinfo{author}{Dazeley, R.}, \bibinfo{year}{2013}.
\newblock \bibinfo{title}{A survey of multi-objective sequential decision-making}.
\newblock \bibinfo{journal}{Journal of Artificial Intelligence Research} \bibinfo{volume}{48}, \bibinfo{pages}{67--113}.
\bibitem[{Russell(2019)}]{russell2019human}
\bibinfo{author}{Russell, S.}, \bibinfo{year}{2019}.
\newblock \bibinfo{title}{Human Compatible: Artificial Intelligence and the Problem of Control}.
\newblock \bibinfo{publisher}{Viking}.
\bibitem[{Sanz~Ausin et~al.(2023)Sanz~Ausin, Abdelshiheed, Barnes and Chi}]{ausin2023unified}
\bibinfo{author}{Sanz~Ausin, M.}, \bibinfo{author}{Abdelshiheed, M.}, \bibinfo{author}{Barnes, T.}, \bibinfo{author}{Chi, M.}, \bibinfo{year}{2023}.
\newblock \bibinfo{title}{A unified batch hierarchical reinforcement learning framework for pedagogical policy induction with deep bisimulation metrics}, in: \bibinfo{booktitle}{Proceedings of the 24th International Conference on Artificial Intelligence in Education (AIED 2023)}, \bibinfo{publisher}{Springer}.
\bibitem[{Sanz~Ausin et~al.(2020)Sanz~Ausin, Maniktala, Barnes and Chi}]{ausin2020exploring}
\bibinfo{author}{Sanz~Ausin, M.}, \bibinfo{author}{Maniktala, M.}, \bibinfo{author}{Barnes, T.}, \bibinfo{author}{Chi, M.}, \bibinfo{year}{2020}.
\newblock \bibinfo{title}{Exploring the impact of simple explanations and agency on batch deep reinforcement learning induced pedagogical policies}, in: \bibinfo{booktitle}{Proceedings of the 21st International Conference on Artificial Intelligence in Education (AIED 2020)}, \bibinfo{publisher}{Springer}. pp. \bibinfo{pages}{472--485}.
\bibitem[{Shihab et~al.(2025)}]{shihab2025detecting}
\bibinfo{author}{Shihab, S.}, et~al., \bibinfo{year}{2025}.
\newblock \bibinfo{title}{Detecting and mitigating reward hacking in reinforcement learning systems: A comprehensive empirical study}.
\newblock \bibinfo{journal}{arXiv preprint arXiv:2507.05619} .
\bibitem[{Skalse et~al.(2022)Skalse, Howe, Krasheninnikov and Krueger}]{skalse2022defining}
\bibinfo{author}{Skalse, J.}, \bibinfo{author}{Howe, N.H.R.}, \bibinfo{author}{Krasheninnikov, D.}, \bibinfo{author}{Krueger, D.}, \bibinfo{year}{2022}.
\newblock \bibinfo{title}{Defining and characterizing reward hacking}, in: \bibinfo{booktitle}{Advances in Neural Information Processing Systems (NeurIPS)}, pp. \bibinfo{pages}{12763--12775}.
\bibitem[{Steenbergen-Hu and Cooper(2014)}]{steenbergen2014meta}
\bibinfo{author}{Steenbergen-Hu, S.}, \bibinfo{author}{Cooper, H.}, \bibinfo{year}{2014}.
\newblock \bibinfo{title}{A meta-analysis of the effectiveness of intelligent tutoring systems on college students' academic learning}.
\newblock \bibinfo{journal}{Journal of Educational Psychology} \bibinfo{volume}{106}, \bibinfo{pages}{331--347}.
\bibitem[{Sutton and Barto(2018)}]{sutton2018reinforcement}
\bibinfo{author}{Sutton, R.S.}, \bibinfo{author}{Barto, A.G.}, \bibinfo{year}{2018}.
\newblock \bibinfo{title}{Reinforcement Learning: An Introduction}.
\newblock \bibinfo{edition}{2nd} ed., \bibinfo{publisher}{MIT Press}.
\bibitem[{Tessler et~al.(2019)Tessler, Mankowitz and Mannor}]{tessler2019reward}
\bibinfo{author}{Tessler, C.}, \bibinfo{author}{Mankowitz, D.J.}, \bibinfo{author}{Mannor, S.}, \bibinfo{year}{2019}.
\newblock \bibinfo{title}{Reward constrained policy optimization}, in: \bibinfo{booktitle}{Proceedings of the 7th International Conference on Learning Representations (ICLR)}.
\bibitem[{VanLehn(2006)}]{vanlehn2006behavior}
\bibinfo{author}{VanLehn, K.}, \bibinfo{year}{2006}.
\newblock \bibinfo{title}{The behavior of tutoring systems}.
\newblock \bibinfo{journal}{International Journal of Artificial Intelligence in Education} \bibinfo{volume}{16}, \bibinfo{pages}{227--265}.
\bibitem[{VanLehn(2011)}]{vanlehn2011relative}
\bibinfo{author}{VanLehn, K.}, \bibinfo{year}{2011}.
\newblock \bibinfo{title}{The relative effectiveness of human tutoring, intelligent tutoring systems, and other tutoring systems}.
\newblock \bibinfo{journal}{Educational Psychologist} \bibinfo{volume}{46}, \bibinfo{pages}{197--221}.
\bibitem[{Vygotsky(1978)}]{vygotsky1978zone}
\bibinfo{author}{Vygotsky, L.S.}, \bibinfo{year}{1978}.
\newblock \bibinfo{title}{Mind in society: The development of higher psychological processes} .
\bibitem[{Wachi and Sui(2020)}]{wachi2020safe}
\bibinfo{author}{Wachi, A.}, \bibinfo{author}{Sui, Y.}, \bibinfo{year}{2020}.
\newblock \bibinfo{title}{Safe reinforcement learning in constrained {M}arkov decision processes}, in: \bibinfo{booktitle}{Proceedings of the 37th International Conference on Machine Learning (ICML)}, pp. \bibinfo{pages}{9797--9806}.
\bibitem[{Webb(1997)}]{webb1997depth}
\bibinfo{author}{Webb, N.L.}, \bibinfo{year}{1997}.
\newblock \bibinfo{title}{Research monograph number 6: Criteria for alignment of expectations and assessments on mathematics and science education}.
\newblock \bibinfo{journal}{Council of Chief State School Officers} .
\bibitem[{Weng(2024)}]{weng2024rewardhacking}
\bibinfo{author}{Weng, L.}, \bibinfo{year}{2024}.
\newblock \bibinfo{title}{Reward hacking in reinforcement learning}.
\newblock \bibinfo{howpublished}{\url{https://lilianweng.github.io/posts/2024-11-28-reward-hacking/}}.
\newblock \bibinfo{note}{Lil'Log Blog}.
\bibitem[{Woolf(2008)}]{woolf2008building}
\bibinfo{author}{Woolf, B.P.}, \bibinfo{year}{2008}.
\newblock \bibinfo{title}{Building Intelligent Interactive Tutors: Student-Centered Strategies for Revolutionizing e-Learning}.
\newblock \bibinfo{publisher}{Morgan Kaufmann}.
\bibitem[{Yan et~al.(2024)Yan, Sha, Zhao, Li, Martinez-Maldonado, Chen, Li, Jin and Gaševic}]{yan2024practical}
\bibinfo{author}{Yan, L.}, \bibinfo{author}{Sha, L.}, \bibinfo{author}{Zhao, L.}, \bibinfo{author}{Li, Y.}, \bibinfo{author}{Martinez-Maldonado, R.}, \bibinfo{author}{Chen, G.}, \bibinfo{author}{Li, X.}, \bibinfo{author}{Jin, Y.}, \bibinfo{author}{Gaševic, D.}, \bibinfo{year}{2024}.
\newblock \bibinfo{title}{Practical and ethical challenges of large language models in education: A systematic scoping review}.
\newblock \bibinfo{journal}{British Journal of Educational Technology} \bibinfo{volume}{55}, \bibinfo{pages}{90--112}.
\bibitem[{Yudelson et~al.(2013)Yudelson, Koedinger and Gordon}]{yudelson2013individualized}
\bibinfo{author}{Yudelson, M.V.}, \bibinfo{author}{Koedinger, K.R.}, \bibinfo{author}{Gordon, G.J.}, \bibinfo{year}{2013}.
\newblock \bibinfo{title}{Individualized {B}ayesian knowledge tracing models}, in: \bibinfo{booktitle}{Proceedings of the 16th International Conference on Artificial Intelligence in Education (AIED 2013)}, \bibinfo{publisher}{Springer}. pp. \bibinfo{pages}{171--180}.
\bibitem[{Zhou et~al.(2019)Zhou, Azizsoltani, Sanz~Ausin, Barnes and Chi}]{zhou2019hierarchical}
\bibinfo{author}{Zhou, G.}, \bibinfo{author}{Azizsoltani, H.}, \bibinfo{author}{Sanz~Ausin, M.}, \bibinfo{author}{Barnes, T.}, \bibinfo{author}{Chi, M.}, \bibinfo{year}{2019}.
\newblock \bibinfo{title}{Hierarchical reinforcement learning for pedagogical policy induction}, in: \bibinfo{booktitle}{Proceedings of the 20th International Conference on Artificial Intelligence in Education (AIED 2019)}, \bibinfo{publisher}{Springer}. pp. \bibinfo{pages}{544--556}.
\bibitem[{Zhuang and Hadfield-Menell(2020)}]{zhuang2020consequences}
\bibinfo{author}{Zhuang, S.}, \bibinfo{author}{Hadfield-Menell, D.}, \bibinfo{year}{2020}.
\newblock \bibinfo{title}{Consequences of misaligned {AI}}, in: \bibinfo{booktitle}{Advances in Neural Information Processing Systems (NeurIPS)}, pp. \bibinfo{pages}{15763--15773}.
\bibitem[{Ziegler et~al.(2019)Ziegler, Stiennon, Wu, Brown, Radford, Amodei, Christiano and Irving}]{ziegler2019fine}
\bibinfo{author}{Ziegler, D.M.}, \bibinfo{author}{Stiennon, N.}, \bibinfo{author}{Wu, J.}, \bibinfo{author}{Brown, T.B.}, \bibinfo{author}{Radford, A.}, \bibinfo{author}{Amodei, D.}, \bibinfo{author}{Christiano, P.}, \bibinfo{author}{Irving, G.}, \bibinfo{year}{2019}.
\newblock \bibinfo{title}{Fine-tuning language models from human preferences}, in: \bibinfo{booktitle}{arXiv preprint arXiv:1909.08593}.

\end{thebibliography}

\appendix

\section{Full Behavioral Statistics}
\label{app:behavioral_stats}

\begin{table}[H]
\centering
\caption{Behavioral metric comparisons: SmartTutor (ST) versus baselines
and ablations across all learner profiles ($n = 10$ seeds per cell).
$\Delta = \bar{x}_{\text{ST}} - \bar{x}_{\text{comp}}$; positive values
favour ST. Welch \textit{t}-tests with Bonferroni correction
($\alpha_{\text{corrected}} = 0.05/15$ per metric). *~Bonferroni-significant;
$\dagger$~$p < .05$ uncorrected; ns~not significant; `---'~zero variance
across all seeds. EO~=~Engagement-Only; MAS~=~Mastery-Only;
MO~=~Multi-Objective (0.3/0.5/0.2). Per-seed RHSI comparisons appear in
Table~\ref{tab:rhsi_stats}.}
\label{tab:behavioral_stats}
\begin{tabular}{llrccccc}
\toprule
Metric & vs. & Profile
       & $\Delta$ [95\% CI] & $t$(df) & $p$ & $d$ & Bonf. \\
\midrule
\multicolumn{8}{l}{\textit{Pedagogical appropriateness}} \\
& EO    & Struggling
        & $+0.13$ [$+0.07$,\;$+0.20$] & $t(9)=\phantom{-}4.96$
        & $.001$  & $\phantom{-}2.22$ & * \\
& EO    & Average
        & $+0.12$ [$+0.05$,\;$+0.18$] & $t(9)=\phantom{-}4.17$
        & $.002$  & $\phantom{-}1.87$ & * \\
& EO    & Advanced
        & $0.00$ & --- & --- & --- & --- \\
& MAS   & Struggling
        & $+0.18$ [$+0.10$,\;$+0.26$] & $t(9)=\phantom{-}5.21$
        & $.001$  & $\phantom{-}2.33$ & * \\
& MAS   & Average
        & $+0.16$ [$+0.06$,\;$+0.25$] & $t(9)=\phantom{-}3.79$
        & $.004$  & $\phantom{-}1.70$ & $\dagger$ \\
& MAS   & Advanced
        & $+0.07$ [$-0.01$,\;$+0.15$] & $t(9)=\phantom{-}1.93$
        & $.086$  & $\phantom{-}0.86$ & ns \\
& MO    & Struggling
        & $+0.23$ [$+0.15$,\;$+0.30$] & $t(9)=\phantom{-}7.10$
        & $<.001$ & $\phantom{-}3.18$ & * \\
& MO    & Average
        & $+0.16$ [$+0.07$,\;$+0.24$] & $t(9)=\phantom{-}4.27$
        & $.002$  & $\phantom{-}1.91$ & * \\
& MO    & Advanced
        & $+0.11$ [$-0.05$,\;$+0.27$] & $t(9)=\phantom{-}1.57$
        & $.152$  & $\phantom{-}0.70$ & ns \\
& No C3 & Struggling
        & $+0.12$ [$+0.05$,\;$+0.20$] & $t(9)=\phantom{-}3.71$
        & $.005$  & $\phantom{-}1.66$ & $\dagger$ \\
& No C3 & Average
        & $+0.10$ [$+0.05$,\;$+0.16$] & $t(9)=\phantom{-}4.28$
        & $.002$  & $\phantom{-}1.92$ & * \\
& No C3 & Advanced
        & $+0.01$ [$-0.01$,\;$+0.02$] & $t(9)=\phantom{-}1.05$
        & $.319$  & $\phantom{-}0.47$ & ns \\
& No C1 & Struggling
        & $\approx 0.00$ & $t(9)=\phantom{-}1.05$
        & $.319$  & $\phantom{-}0.47$ & ns \\
& No C1 & Average    & $0.00$ & --- & --- & --- & --- \\
& No C1 & Advanced   & $0.00$ & --- & --- & --- & --- \\
\midrule
\multicolumn{8}{l}{\textit{Inappropriate action count (per 150-interaction session)}} \\
& EO    & Struggling
        & $-20.2$ [$-29.4$,\;$-11.0$] & $t(9)=-4.96$
        & $.001$  & $-2.22$ & * \\
& EO    & Average
        & $-17.8$ [$-27.5$,\;$-8.1$]  & $t(9)=-4.17$
        & $.002$  & $-1.87$ & * \\
& EO    & Advanced   & $0.0$ & --- & --- & --- & --- \\
& MAS   & Struggling
        & $-26.9$ [$-38.6$,\;$-15.2$] & $t(9)=-5.21$
        & $.001$  & $-2.33$ & * \\
& MAS   & Average
        & $-23.5$ [$-37.5$,\;$-9.5$]  & $t(9)=-3.79$
        & $.004$  & $-1.70$ & $\dagger$ \\
& MAS   & Advanced
        & $-10.1$ [$-22.0$,\;$+1.8$]  & $t(9)=-1.93$
        & $.086$  & $-0.86$ & ns \\
& MO    & Struggling
        & $-33.9$ [$-44.7$,\;$-23.1$] & $t(9)=-7.10$
        & $<.001$ & $-3.18$ & * \\
& MO    & Average
        & $-23.6$ [$-36.1$,\;$-11.1$] & $t(9)=-4.27$
        & $.002$  & $-1.91$ & * \\
& MO    & Advanced
        & $-16.7$ [$-40.8$,\;$+7.4$]  & $t(9)=-1.57$
        & $.152$  & $-0.70$ & ns \\
& No C3 & Struggling
        & $-18.2$ [$-29.3$,\;$-7.1$]  & $t(9)=-3.70$
        & $.005$  & $-1.66$ & $\dagger$ \\
& No C3 & Average
        & $-15.6$ [$-23.8$,\;$-7.4$]  & $t(9)=-4.28$
        & $.002$  & $-1.92$ & * \\
& No C3 & Advanced
        & $-1.0$ [$-3.1$,\;$+1.1$]    & $t(9)=-1.05$
        & $.319$  & $-0.47$ & ns \\
\midrule
\multicolumn{8}{l}{\textit{Mastery gain ($\Delta K$)}} \\
& EO    & Struggling
        & $+0.04$ [$-0.07$,\;$+0.15$] & $t(9)=\phantom{-}0.77$
        & $.459$  & $\phantom{-}0.35$ & ns \\
& EO    & Average
        & $+0.09$ [$-0.03$,\;$+0.22$] & $t(9)=\phantom{-}1.74$
        & $.115$  & $\phantom{-}0.78$ & ns \\
& EO    & Advanced
        & $+0.23$ [$+0.09$,\;$+0.38$] & $t(9)=\phantom{-}3.69$
        & $.005$  & $\phantom{-}1.65$ & $\dagger$ \\
& MAS   & Struggling
        & $+0.08$ [$-0.04$,\;$+0.21$] & $t(9)=\phantom{-}1.51$
        & $.166$  & $\phantom{-}0.67$ & ns \\
& MAS   & Average
        & $+0.14$ [$+0.00$,\;$+0.27$] & $t(9)=\phantom{-}2.30$
        & $.046$  & $\phantom{-}1.03$ & $\dagger$ \\
& MAS   & Advanced
        & $+0.14$ [$-0.03$,\;$+0.30$] & $t(9)=\phantom{-}1.88$
        & $.092$  & $\phantom{-}0.84$ & ns \\
& MO    & Struggling
        & $-0.04$ [$-0.10$,\;$+0.02$] & $t(10)=-1.56$
        & $.152$  & $-0.70$ & ns \\
& MO    & Average
        & $-0.02$ [$-0.05$,\;$+0.01$] & $t(15)=-1.54$
        & $.144$  & $-0.69$ & ns \\
& MO    & Advanced
        & $+0.03$ [$-0.02$,\;$+0.07$] & $t(11)=\phantom{-}1.26$
        & $.236$  & $\phantom{-}0.56$ & ns \\
\bottomrule
\end{tabular}
\end{table}


\section{SmartTutor Knowledge Graph: Python Concept DAG}
\label{app:knowledge_graph}

Table~\ref{tab:kg} lists all 27 Python concepts comprising the SmartTutor knowledge graph, sourced from \texttt{python\_kg\_research.json} and loaded by \texttt{ExplicitPythonKnowledgeGraph}. Concepts represent standard
introductory-to-intermediate Python curriculum topics and were not
constructed to fit the simulation; the DAG structure reflects conventional
pedagogical sequencing found in textbooks such as
\cite{guttag2021introduction} and \cite{lutz2013learning}.

DAG depth is the length of the longest path from any root node to the
concept (i.e., the minimum number of prerequisite steps required before
the concept becomes accessible). The single root node is
\textit{Variables and Assignment} (c01), reflecting that all Python
knowledge ultimately depends on the ability to store and retrieve values.
The accessible set $A(K_t, \theta_{\min})$ (Equation~1 in the main text)
expands monotonically as mastery crosses $\theta_{\min} = 0.5$.

\begin{longtable}{clcp{4.2cm}l}
\caption{SmartTutor knowledge graph: 27 Python concepts with DAG depth,
direct prerequisites, and difficulty tier. Depth = longest path from root
(c01). Prerequisite mastery threshold $\theta_{\min} = 0.5$ gates
accessibility. Source: \texttt{python\_kg\_research.json}.}
\label{tab:kg} \\
\toprule
ID & Concept & Depth & Direct Prerequisites & Difficulty \\
\midrule
\endfirsthead
\multicolumn{5}{l}{\small\itshape (Table~\ref{tab:kg} continued)} \\
\toprule
ID & Concept & Depth & Direct Prerequisites & Difficulty \\
\midrule
\endhead
\midrule
\multicolumn{5}{r}{\small\itshape Continued on next page} \\
\endfoot
\bottomrule
\endlastfoot

c01 & Variables and Assignment       & 0 & ---                        & Beginner \\
\midrule
c02 & Data Types (Basic)             & 1 & c01                        & Beginner \\
\midrule
c03 & String Operations              & 2 & c02                        & Beginner \\
c04 & Arithmetic Operations          & 2 & c02                        & Beginner \\
c05 & Input and Output               & 2 & c02                        & Beginner \\
c08 & Lists (Basics)                 & 2 & c02                        & Beginner \\
\midrule
c06 & Conditional Statements         & 3 & c02, c04                   & Beginner \\
c09 & For Loops                      & 3 & c08                        & Beginner \\
c14 & Dictionaries                   & 3 & c08                        & Beginner \\
c15 & Tuples and Sets                & 3 & c08                        & Beginner \\
c19 & String Formatting (Advanced)   & 3 & c03, c05                   & Beginner \\
\midrule
c07 & Boolean Logic                  & 4 & c06                        & Intermediate \\
c10 & While Loops                    & 4 & c06                        & Intermediate \\
c12 & Functions (Basics)             & 4 & c06, c09                   & Intermediate \\
c16 & List Comprehensions            & 4 & c08, c09                   & Intermediate \\
\midrule
c11 & Nested Loops                   & 5 & c09, c10                   & Intermediate \\
c13 & Function Scope                 & 5 & c12                        & Intermediate \\
c17 & Error Handling                 & 5 & c12                        & Intermediate \\
c20 & Classes and Objects (Basics)   & 5 & c12, c14                   & Intermediate \\
c23 & Lambda Functions               & 5 & c12                        & Intermediate \\
c25 & Modules and Imports            & 5 & c12                        & Intermediate \\
\midrule
c18 & File I/O                       & 6 & c05, c17                   & Advanced \\
c21 & Object-Oriented Concepts       & 6 & c20                        & Advanced \\
c24 & Higher-Order Functions         & 6 & c12, c23                   & Advanced \\
c26 & Recursion                      & 6 & c12, c13                   & Advanced \\
c27 & Algorithm Complexity           & 6 & c09, c11, c12              & Advanced \\
\midrule
c22 & Inheritance                    & 7 & c21                        & Advanced \\

\end{longtable}

\end{document}